%% file: main.tex
\documentclass[runningheads]{llncs}

\usepackage[mobile]{eccv}

\usepackage{eccvabbrv}

\usepackage{graphicx}
\usepackage{booktabs}

\usepackage[accsupp]{axessibility}  %
\usepackage{multirow}

\usepackage{tabularray}
\usepackage{times}
\usepackage{epsfig}
\usepackage{graphicx}
\usepackage{amsmath}
\usepackage{amssymb}
\usepackage{array}
\usepackage{multirow}
\usepackage{booktabs,threeparttable}
\usepackage{float}
\usepackage{wrapfig}
\usepackage{colortbl}
\usepackage{subcaption}
\input{preamble}
\usepackage{soul}

\usepackage[pagebackref,breaklinks,colorlinks,citecolor=eccvblue]{hyperref}

\usepackage{orcidlink}

\newcommand\blfootnote[1]{%
  \begingroup
  \renewcommand\thefootnote{}\footnote{#1}%
  \addtocounter{footnote}{-1}%
  \endgroup
}

\begin{document}

\title{TAM-VT: Transformation-Aware Multi-scale Video Transformer for Segmentation and Tracking}

\titlerunning{TAM-VT: Transformation-Aware Multi-scale Semi-VOS}

\author{
Raghav Goyal$^{*1,2}$ \hspace{0.25in} Wan-Cyuan Fan$^{*1,2}$ \hspace{0.25in} Mennatullah Siam$^{4}$ \\ Leonid Sigal$^{1,2,3}$ 
}

\authorrunning{Goyal and Fan et al.}

\institute{$^1$University of British Columbia \hspace{0.2in}
$^2$Vector Institute for AI \hspace{0.2in}
$^3$CIFAR AI Chair\\
$^4$Ontario Tech University
}

\maketitle

\input{sec/0_abstract}

\input{sec/1_introduction}

\input{sec/2_related_work}

\input{sec/3_approach}

\input{sec/4_experiments}

\input{sec/5_conclusion}

\section*{Acknowledgements}
{\small This work was funded, in part, by the Vector Institute for AI, Canada CIFAR AI Chairs, NSERC CRC, and NSERC DGs. Resources used in preparing this research were provided, in part, by the Province of Ontario, the Government of Canada through CIFAR, the Digital Research Alliance of Canada\footnote{\url{alliance.can.ca}}, companies\footnote{\url{https://vectorinstitute.ai/\#partners}} sponsoring the Vector Institute, and Advanced Research Computing at the University of British Columbia. Additional hardware support was provided by John R. Evans Leaders Fund CFI grant and Compute Canada under the Resource Allocation Competition award.}

\bibliographystyle{splncs04}
\bibliography{main}

\input{sec/suppl}

\end{document}

%% file: preamble.tex
\usepackage[dvipsnames]{xcolor}

\newcommand{\todo}[1]{{\color{red}#1}}

\newcommand{\our}{our model }
\newcommand{\ourmodel}{TAM-VT}

%% file: sec/0_abstract.tex
\begin{abstract}
Video Object Segmentation (VOS) has
emerged as an increasingly important problem with availability of larger datasets and more complex and realistic settings, which involve long videos with global motion (\eg, in egocentric settings), depicting small objects undergoing both rigid and non-rigid (including state) deformations. While a number of recent approaches have been explored for this task, these data characteristics still present challenges. In this work we propose a novel, clip-based DETR-style encoder-decoder architecture, which focuses on systematically analyzing and addressing aforementioned challenges.
Specifically, we propose a novel transformation-aware loss that focuses learning on portions of the video where an object undergoes significant deformations -- a form of ``soft" hard examples mining. 
Further, we propose a multiplicative time-coded memory, beyond vanilla additive positional encoding, which helps propagate context across long videos. Finally, we incorporate these in our proposed holistic multi-scale video transformer for tracking via multi-scale memory matching and decoding to ensure sensitivity and accuracy for long videos and small objects.
Our model enables on-line inference with long videos in a windowed fashion, by breaking the video into clips and propagating context among them.
We illustrate that short clip length and longer memory with learned time-coding are important design choices for improved %
performance. 
Collectively, these technical contributions enable our model to achieve new state-of-the-art (SoTA) performance on two complex egocentric datasets -- VISOR \cite{darkhalil2022epic} and VOST \cite{tokmakov2023breaking}, while achieving comparable to SoTA results on the conventional VOS benchmark, DAVIS'17~\cite{pont20172017}. A series of detailed ablations validate our design choices as well as provide insights into the importance of parameter choices and their impact on performance.  
\keywords{Video Object Segmentation \and Transformation-aware loss \and Holistic multi-scale video transformer \and Long videos \and Small objects \and Egocentric datasets}

\blfootnote{* equal contribution.}

\end{abstract}

%% file: sec/1_introduction.tex
\vspace{-0.2in}
\input{figures/teaser}
\section{Introduction}
Semi-automatic Video Object Segmentation (Semi-VOS) involves segmenting an object of interest in a video sequence given the provided initial mask in the first frame. 
This task has attracted significant attention 
in the literature~\cite{zhou2022survey} and is both interesting and challenging, 
involving aspects of detection/segmentation and temporal tracking. 

Previous work on Semi-VOS often leverages memory matching, first proposed by Space-Time Memory networks (STM)~\cite{oh2019video}. The idea is simple, to encode the initial frame-mask pair (and, possibly later, frames and their inferred masks, as pairs) in neural memory, in order to perform spatiotemporal matching that helps mask prediction in new frame(s). A recent development focused on performing per-clip inference while utilizing this spatiotemporal memory matching instead of per-frame inference~\cite{park2022per}.
However, such approaches are challenged when objects are small ({\em i.e.}, object size is small compared to patches used for matching) or when objects undergo significant deformations. There has been previous works exploring multi-scale processing in Semi-VOS~\cite{seong2021hierarchical,yang2021collaborative}. However, they focused their multi-scale effort in transformers 
on memory matching only.

Recently, there has been a rise in the Semi-VOS datasets dedicated to egocentric videos which focus on such complex deformations and long videos~\cite{tokmakov2023breaking,darkhalil2022epic}. 
In these %
datasets, objects can undergo extreme state changes, {\em e.g.}, a banana being peeled/cut;  or objects being molded into different shapes. 
Previous Semi-VOS approaches, especially memory-based ones, tend to face difficulty in such complex deformation scenarios~\cite{tokmakov2023breaking}. Recent memory-based techniques that handled long videos~\cite{cheng2022xmem,hong2023lvos} or were able to associate multiple objects without adhoc post-processing~\cite{yang2021associating} have shown relatively better performance. Nonetheless, current 
methods are not equipped with mechanisms specifically designed for these extreme state changes or small objects.

In this paper, we focus on addressing the aforementioned challenges by proposing a novel transformation-aware loss that is designed to mine hard examples, arising from abrupt object state changes, using a reweighing scheme. Additionally, we propose a holistic multi-scale video transformer for Semi-VOS which incorporates multi-scale memory matching and decoding within a DETR-style framework. We refer to it as holistic, since it unifies multi-scale processing in both memory matching and decoding within our video transformer unlike previous methods \cite{seong2021hierarchical,yang2021collaborative}. Our method features a clip-based windowed approach that excels in capturing objects in long videos. Illustration of our approach and improvements it achieves are shown in~\cref{fig:intro}.
The proposed multi-scale framework allows better local matching, which has significant impact on the ability to deal with small objects or objects undergoing complex deformations.
Further, the learned multiplicative time-coding, within this memory, allows our model to leverage recency (nearby frames tend to be more relevant when segmenting a current one). Collectively, these design choices lead to SoTA performance on two challenging egocentric datasets -- VISOR \cite{darkhalil2022epic} and VOST \cite{tokmakov2023breaking}, while showing comparable results on conventional VOS datasets.

Our contributions can be summarized as follows: 
(1) We propose a novel multi-scale memory matching and decoding unified in a DETR-style clip-based framework for Semi-VOS, capable of dealing with arbitrary long videos in a windowed fashion. As part of this framework, we (2) introduce a novel multi-scale multiplicative time-coded memory, designed to be particularly effective for objects that are small or can undergo complex deformations. We further, (3) introduce transformation-aware training objective that guides the model to focus learning on portions of the video where an object is transformed, thereby enhancing its ability to track objects across such changes. Finally, (4) through a series of detailed ablations we validate our design as well as provide insights into the importance of parameter choices ({\em e.g.}, shorter clip and longer memory length) and their impact on performance.

Experimental results demonstrate superior performance of our approach compared with previous state-of-the-art Semi-VOS techniques, resulting in $\approx 7\%$ improvement in VISOR \cite{darkhalil2022epic} and $\approx 1\%$ in VOST \cite{tokmakov2023breaking} for all metrics. Especially, on longer videos and small objects of VOST, we outperform the next best method by $7\%$ and $4\%$, resp, while performing competitively on the conventional VOS benchmark, DAVIS'17~\cite{pont20172017}.

%% file: figures/teaser.tex
\begin{figure}[t]
  \centering
  \includegraphics[page=3,trim={0 650 50 0}, clip, width=0.9\columnwidth]{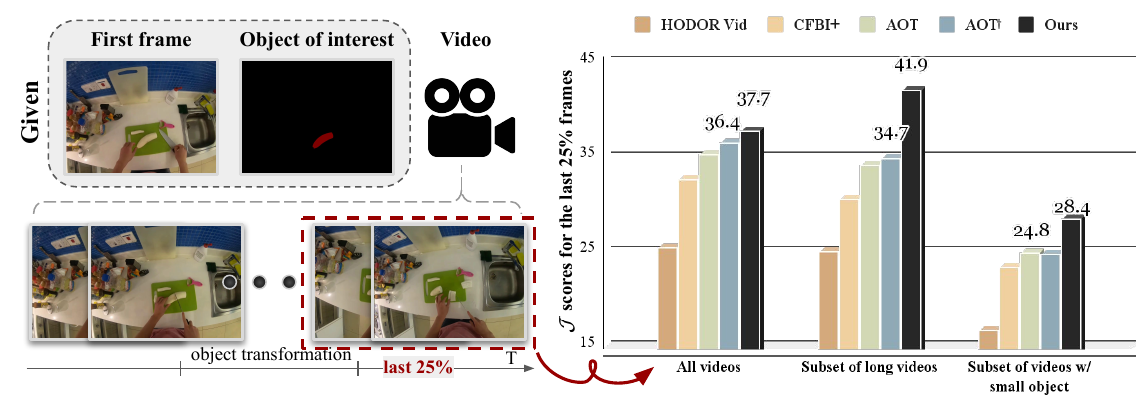}
  \vspace{-4mm}
  \caption{{\bf Illustration of Our Approach.} Given the first frame and the mask for the object of interest, our model adeptly tracks the object through the video via producing a sequence of segmentation masks. Despite potential transformations of the object, our approach achieves better performance on long videos (\textgreater 20 sec) and videos with small object (\textless 0.5\% frame area). $\dag$ denotes results reproduced by us using official code.} %
  \label{fig:intro}
  \vspace{-6mm}
\end{figure}

%% file: sec/2_related_work.tex
\section{Related Works}

\noindent
{\bf Video Object Segmentation (VOS).}
Video object segmentation separates foreground object(s) in a video from the background in a class agnostic manner. 
This task has two sub-categories~\cite{zhou2022survey}: {\em automatic} and {\em semi-automatic}. 
The former defines foreground object(s) based on their saliency in terms of appearance and motion, while the latter defines them based on tracking manually initialized masks in the first frame of the video. 
In this work, we focus on \textit{semi-automatic} formulation. 
Note, previously, this task was referred to as \textit{semi-supervised}~\cite{perazzi2016benchmark}, which we avoid to reduce confusion with techniques that use additional unlabelled data~\cite{zhou2022survey}. We refer to our closest work from the automatic video object segmentation literature that devised a unified multi-scale encoder-decoder video transformer~\cite{karim2023med}. However, their work was not designed for tracking, as such they do not incorporate multi-scale memory matching nor do they handle abrupt state changes which is necessary in tracking, unlike our devised novel loss and framework.

\vspace{0.08in}
\noindent
{\bf Semi-automatic VOS (Semi-VOS).}
Semi-automatic methods perform segmentation and tracking with various challenges in terms of occlusions, objects deformation, motion blur, tracking of small objects and differentiating similar ones. Recently, there are some benchmarks proposed on egocentric videos~\cite{darkhalil2022epic,tokmakov2023breaking,yu2023video} with few focusing on objects undergoing extreme state changes, {\em e.g.}, eggs being broken. Tracking objects under these extreme changes, where shape, consistency, colour and texture of the object drastically change, presents additional challenges. 

Semi-VOS approaches can be categorized into {\em online fine-tuning} based~\cite{xiao2018monet,yang2018efficient,voigtlaender2017online,caelles2017one}, {\em propagation} based~\cite{jang2017online,jampani2017video,bao2018cnn} and {\em matching} based~\cite{hu2018videomatch,voigtlaender2019feelvos,oh2019video,yang2021associating,cheng2023putting} methods. 
Matching based methods, usually relying on memory-based techniques, have proven to be most effective and robust~\cite{oh2019video,yang2021associating,park2022per,cheng2022xmem}.
We build on this line of works, but propose an approach that is specifically designed 
to accommodate the aforementioned challenges.

\vspace{0.08in}
\noindent
{\bf Role of memory in VOS. } 
Space-Time Memory (STM)~\cite{oh2019video} proposed a memory-based technique that spawned  multiple consequent approaches~\cite{yang2021associating,cheng2022xmem,park2022per,hong2023lvos}. AOT~\cite{yang2021associating} extended memory reading to track multiple objects simultaneously, while incorporating both global and local attention mechanisms. XMem~\cite{cheng2022xmem} extended the use of memory for long-video sequences by creating memory modules at three different time-levels: long-term, short-term and per-frame. While previous methods focused on per-frame inference, PCVOS~\cite{park2022per} proposed inference and updates to memory on a clip-level. Overall, these methods focused on matching with the memory bank on a single-spatial scale. However, in this work, we propose multi-scale memory matching and decoding which is particularly effective for small objects. Similar to PCVOS~\cite{park2022per}, we use clip-based memory updates, but in addition we propose a relative time-encoding for memory to learn recency, particularly useful for long-videos. Compared to XMem~\cite{cheng2022xmem}, our time-encoding is learned on per-frame level instead of three explicit time-levels.
Finally, unlike previous works we propose a transformation-aware loss that takes state changes into account as a form of hard example mining during train.

%% file: sec/3_approach.tex
\vspace{-0.05in}
\section{Approach}
\label{sec:approach}

Given a video, a reference frame and the corresponding ground-truth segmentation mask indicating the object of interest, our goal is to predict the sequence of segmentation masks of the object for all the frames in the video. 

\vspace{-0.1in}
\subsection{Overview}
\vspace{-0.05in}
Our model builds on DETR-style video encoder-decoder architectures \cite{carion2020end, yang2022tubedetr,cheng2022masked}. Our motivation is to track objects through challenging scenarios ({\em e.g.}, tracking small objects) and under extreme state changes across long videos. 
Therefore we endow the aforementioned architecture with 
three key components; (i) multi-scale matching encoder, (ii) clip-based memory, and (iii) multi-scale decoder.

\input{figures/framework}

Given an input video and ground-truth segmentation mask of a desired object, we make predictions by sub-dividing the video into non-overlapping clips, and processing the resulting clips in a sequential order while propagating information through the use of the clip-based memory.

Fig. \ref{fig:Architecture_1} shows an overview of our approach. For each query clip, we encode frames in the clip using a 2D-CNN visual backbone, yielding frame-features (or tokens) at multiple-scales or feature hierarchies. Mask predictions of previous video clips or ground-truth, %
for the first frame, are taken as memory. We perform dense token matching between frame features of the query clip and memory at multiple-scales using our \textit{Multi-Scale Matching Encoder} to obtain similarity between the query and past clips. In particular, we use a multiplicative \textit{Relative Time Encoding (RTE)} for memory to encode relative time difference of past clips w.r.t. the current query clip. Thus, our approach enables more accurate long temporal context modelling. We use the resulting similarity to obtain encoded mask features for the query clip as a weighted combination of the memory mask features. Such multi-scale matching enables propagating object mask from memory, which is especially useful to track small objects. We then utilize our \textit{Multi-Scale Decoder} to obtain the final mask predictions. Specifically, the feature-pyramid is introduced to aggregate features by incorporating the query clip frame-features with encoded mask features in a multi-scale fashion, and then the space-time decoder decodes mask predictions from feature pyramid for the query clip. We then update the memory with the query clip frame features and mask predictions to be used for the next clip. During training, we use \textit{Transformation-aware Reweighting} to re-weigh clip segmentation loss \cite{carion2020end} over time. This can be interpreted as a form of ``soft'' hard examples mining, to account for object transformations which are tightly coupled to the change in relative area, position and number of connected parts of the ground-truth object. We discuss details of our proposed modules in the following sections.

\subsection{Multi-scale Matching Encoder}
\label{sec:multiscale_matching_encoder}
For a query video clip of length $L$, we encode each frame independently using a ResNet \cite{he2016deep} to obtain frame features $X^{q} = \{\mathbf{x}^{q}_{s}\}_{s \leq S}$, where $\mathbf{x}^{q}_{s} \in \mathbb{R}^{L \times h_s \times w_s\times d_{s}} $ is the feature map of query clip at scale $s$. The total number of scales are denoted by $S$, and $h_s$, $w_s$ and $d_{s}$ are height, width and channels of the feature map at scale $s$.

Inspired by previous work \cite{kim2023universal}, originally designed for few-shot tasks and operating on single images, we propose a clip based multi-scale query-memory matching module for dense prediction tasks. We compute similarity between frame features of query and memory at multiple-scales, and obtain mask features for the query clip as a weighted combination of the memory mask features. We define a memory with a maximum size of $N$, and the elements stored in the memory are denoted as $\{X^M, Y^M\}$ , where $X^M = \{\mathbf{x}^{m}_{s}\}_{s \leq S}$, $Y^M_i = \{\mathbf{y}^{m}_{s}\}_{s \leq S}$ are frame and mask features respectively in memory, s.t. $\mathbf{x}^{m}_{s},\mathbf{y}^{m}_{s} \in \mathbb{R}^{n \times h_s \times w_s\times d_{s}}$, with a total size of $n$ ($n \leq N$). The memory contains features of past video clip frames and predicted masks (along with the reference ground-truth frame). We describe update operations applied to memory in \cref{sec:clip-based-memory}.

For each scale $s$, we use multi-head attention layer (MHA) for token matching~\cite{vaswani2017attention,kim2023universal}, where we get tokens for query frames $\{\mathbf{q}_{s}\}_{s \leq S}$ and memory $\{\mathbf{k}_{s}, \mathbf{v}_{s}\}_{s \leq S}$ by applying linear projection and adding 2D positional encoding to query frame features $X^{q}$, and memory frame and mask features $\{X^M, Y^M\}$ respectively, where $\mathbf{q}_{s} \in \mathbb{R}^{L h_s w_s \times d}$ and $\mathbf{k}_{s}, \mathbf{v}_{s} \in \mathbb{R}^{n h_s w_s \times d}$ after flattening the tokens, and $d$ is hidden dimension. The query clip's encoded mask tokens are obtained for each scale $s$ using,

\vspace{-0.1in}
\begin{equation}
    \mathbf{y}^{q,enc}_{s} = 
    \text{MHA}(\mathbf{q}_{s}, \mathbf{k}_{s}, \mathbf{v}_{s}) = \left[\mathbf{o}_1, ..., \mathbf{o}_H \right] w_s^{O},
\end{equation}

\vspace{-0.1in}
\begin{equation}
    \mathbf{o}_h = {\tt Softmax} \left(
    \frac{\mathbf{q}_{s} w_{h, s}^{Q} (\mathbf{k}_{s} w_{h, s}^{K})^{T} }{\sqrt{d_H}}
    \right) \mathbf{v}_{s}w_{h, s}^{V},
    \label{eq:similarity}
\end{equation}

\noindent
where $H$ is number of heads, $d_H$ is the channels, $w^Q_{h,s}, w^K_{h,s}, w^V_{h,s} \in \mathbb{R}^{d \times d_H}$ and $w_s^{O} \in \mathbb{R}^{H d_H \times d}$ for scale $s$ and $\left[.\right]$ is the concatenation operator. After reshaping the encoded mask tokens, we obtain the query clip's encoded mask features denoted as $Y^{q,enc} = \{\mathbf{y}^{q,enc}_{s}\}_{s \leq S}$, where $\mathbf{y}^{q,enc}_{s} \in \mathbb{R}^{L \times h_s \times w_s \times d}$.

\vspace{0.1in}
\noindent
{\bf Relative Time Encoding (RTE).} As described above, the token matching in previous work~\cite{kim2023universal} is designed for static image object segmentation tasks. However, in our setting of video object segmentation, images stored in the memory form a sequence of frames from the given video, and the visual appearance of the foreground object may change significantly, especially in videos with object transformations. 
To address this issue, we propose a multiplicative \textit{Relative-Time Encoding (RTE)} for our matching encoder to 
learn the importance of each frame in the memory when conducting token matching, thereby reducing the potential drift in matching on frames in the memory far from the current timestep.

Specifically, RTE makes the matching dependent on time by modulating similarity weights in \cref{eq:similarity} and helps learn associations based on recency, which is especially useful for extending memory over long-time spans. In particular, based on the memory bank size (=$N$), we define a learnable embedding set as $\{ e_i \in \mathbb{R}^{1 \times i}\}_{i=1}^N$. 
Depending on the number of elements $n$ in the memory, we select the corresponding $e_n$ to modulate the similarity weights. Specifically, we first expand $e_n \in \mathbb{R}^{1\times n}$ to $E_n \in \mathbb{R}^{L h_s w_s \times n h_s w_s}$ by broadcasting the values along the 1-dim and reshaping. And then, we incorporate RTE into \cref{eq:similarity} as follows,

\begin{equation}
    \mathbf{o}_h = {\tt Softmax} \left(
    E_n \circ \frac{\mathbf{q}_{s} w_{h, s}^{Q} (\mathbf{k}_{s} w_{h, s}^{K})^{T} }{\sqrt{d_H}}
    \right) \mathbf{v}_{s}w_{h, s}^{V},
    \label{eq:similarity_RTE}
\end{equation}

\noindent
where $\circ$ represent element-wise multiplication. 
Please refer to Experiments (\cref{sec:exps_analysis}) for more details and ablations to additive relative positional encoding.

\subsection{Multi-scale Decoder}
\vspace{-0.05in}
We obtain feature pyramid (see~\cite{cheng2022masked}) by contexualizing query frame features with the encoded mask features. The motivation is that the use of such high-resolution feature pyramid during decoding would lead to better performance for small objects. 
And finally, we make use of Space-Time Decoder~\cite{yang2022tubedetr} to decode mask predictions from the contextualized feature pyramid for the query clip.

\vspace{0.1in}
\noindent
{\bf Contexualized feature pyramid.} We obtain feature pyramid for the query clip's mask features $Y^{q,fpn}$ by aggregating query clip's image features $X^{q}$ with the encoded multi-scale mask features $Y^{q,enc}$. Specifically, for each scale $s$ in 32, 16, and 8 output stride, %
we add $\mathbf{y}^{q,enc}_{s}$ to contextualize the feature pyramid with the encoded mask features for query clip, as shown in \cref{fig:Architecture_1} \todo{(c)}.

\vspace{-0.1in}
\begin{equation}
    \tilde{\mathbf{y}}^{q, fpn}_{s} =  {\tt Conv2D}(\mathbf{x}^{q}_{s}) + {\tt Upsample}(\mathbf{y}^{q,fpn}_{s-1}) + \mathbf{y}^{q,enc}_{s}
    \label{eq:fpn-first-eq}
\end{equation}
\vspace{-0.1in}
\begin{equation} 
    \mathbf{y}^{q, fpn}_{s} = {\tt Conv2D} \left( \tilde{\mathbf{y}}^{q, fpn}_{s} \right),
\end{equation}

where {\tt Conv2D} and {\tt Upsample} operations are applied frame-wise. Different from conventional FPN \cite{lin2017feature,cheng2022masked}, we aggregate frame features with the encoded mask features (last term in~\cref{eq:fpn-first-eq}) to obtain feature pyramid $Y^{q,fpn} = \{\mathbf{y}^{q,fpn}_{s}\}_{s \leq S}$, where $\mathbf{y}^{q,fpn}_{s} \in \mathbb{R}^{L \times h_s \times w_s \times d}$.

\vspace{0.1in}
\noindent
{\bf Space-Time Decoder.} 
We use learned time embeddings $\{ \mathbf{t}_j \}_{j \leq L} \in \mathbb{R}^{d}$, equal to the length of query clip, and refine them over multiple decoder blocks using the encoded contextualized features $Y^{q,fpn}$ to obtain final mask predictions. In particular, within each decoder block, space-time attention is factorized over time and space by performing temporal self-attention and spatial frame-wise cross-attention operations, interleaved with feed-forward and normalization layers~\cite{yang2022tubedetr}. First, the temporal self-attention allows time embeddings to attend to each other, thereby facilitating temporal interactions across the video clip. Second, the frame-wise cross-attention performs cross-attention with each frame separately, where for a $j^{th}$ frame, the corresponding time embedding $\mathbf{t}_{j}$ cross-attends to its encoded contextualized features $\mathbf{y}^{q,fpn}_{j,s} \in \mathbb{R}^{h_s \times w_s \times d}$. By stacking multiple decoder blocks, the time embeddings gets refined in a hierarchical manner to give $\{ \mathbf{\hat{t}}_j \}_{j \leq L}$, and the final mask prediction is obtained by taking dot-product of $\{ \mathbf{\hat{t}}_j \}_{j \leq L} \in \mathbb{R}^{d}$ with the highest scale (=$S^{th}$) contextualized feature $\{ \mathbf{y}^{q, fpn}_{j,S} \}_{j \leq L} \in \mathbb{R}^{h \times w \times d}$ frame-wise to yield $\hat{Y}^{q} = \{ \mathbf{\hat{y}}^{q}_{j} \}_{j \leq L} \in [0,1]^{h \times w}$, as per-pixel probabilities.

The stacking of multi-scale decoding blocks and the use of multi-scale contextualized features is similar to Mask2Former \cite{cheng2022masked}. Specifically, we use $6$ decoder blocks, where each block receives contextualized features starting from lowest scale $s$ in a round-robin fashion with the appropriate addition of
positional and scale encoding.

Unlike PCVOS \cite{park2022per}, which performs encoding for each frame in query clip independently and performs intra-clip refinement through dense transformer-based attention, we make use of temporal self-attention of time-embeddings to share information within clips in order to account for intra-clip refinement. 

\vspace{-0.1in}
\subsection{Clip-based Memory Module}
\label{sec:clip-based-memory}
We divide the input video into non-overlapping clips of length $L$, and process the video and update memory sequentially one clip at a time. The intuition being that clip-level updates to memory, compared to updates at every frame, trades-off performance with compute time and GPU memory, and at the same time enables long-range modelling \cite{park2022per}. In particular, we perform an update to memory $\{X^M, Y^M\}$ by appending features for frame and predicted mask of the last frame in the query clip (=$L^{th}$ index), {\em i.e.}, $\{X^{q}_{L}, \hat{Y}^{q}_{L}\}$. The memory is implemented as a FIFO (first-in-first-out) queue, which initially gets populated with the features of frame and ground-truth mask of the initial reference frame. We keep the reference frame fixed in memory (as the first index in memory), to have a notion of the initial state of the object if in case there is too much divergence \cite{hong2023lvos}. Lastly, we use the same visual backbone to encode frames and masks.

\vspace{-0.1in}
\subsection{Transformation-aware Reweighting}
\label{sec:transformation_aware_loss}
In our case, objects within a video may undergo transformations, leading to significant changes in their visual appearance and shape. However, treating frames uniformly when forming video-level loss, as done in previous VOS works \cite{yang2021associating,park2022per} could lead to sub-optimal solutions. The reason being that majority of frames contributing to learning are relatively simple, causing redundancy over long-time spans, resulting in failure to track objects of interest post complex transformations.

To address this issue, we frame our objective as an unbalanced learning problem, where the object transformation only occupies a small part of the entire video. We design a reweighting factor to enable the model to place greater emphasis on objects during transformations. Specifically, 
we define our transformation-aware video-level loss as $\mathcal{L}^{\text{tr}}_{\text{total}} = \alpha_1 \mathcal{L}_{\text{dice}} + \alpha_2 \mathcal{L}^{\text{tr}}_{\text{focal}}$, where $\mathcal{L}_{\text{dice}}$ is image-based DICE/F-1 loss~\cite{milletari2016v}, and $\mathcal{L}^{\text{tr}}_{\text{focal}} = \frac{1}{L} \sum^{L}_{t=1} w^t \times l^t_{focal}$, where $l^t_{focal}$ is the focal loss~\cite{lin2017focal} for $t$-th frame, and $w^t$ denotes weight assigned to $t^{th}$ frame. $\alpha_1$ and $\alpha_2$ are hyperparameters to modulate relative importance of loss terms.
Overall, we aim to assign higher weights to frames where the object of interest undergoes transformation. In our fully-supervised setting, we generate weights using ground-truth segmentation masks for the frames in a video, denoted as $M_1, M_2, \dots, M_L$.
Empirically, we found that simply computing the change in area of consecutive ground truth masks is a good indicator to identify frames containing complex transformations. Our observation is that during transformation phase, 
the object gets split/cut, molded, sheared into a different state, resulting in noticeable change in the area of the mask. Based on this, we formulate the weights $\{w^t\}_{t \leq L}$ as,
\setlength{\abovedisplayskip}{2pt}
\begin{align}
    &w^t = \frac{e^{\delta^t/\tau}}{\sum^L_{t=1}e^{\delta^t/\tau}} \times L \text{, where } 
    \\
    &\delta^t = \frac{\| \mathcal{A}(M^{t})- \mathcal{A}(M^{t-1}) \|}{\sum^L_{t=1}\| \mathcal{A}(M^{t})- \mathcal{A}(M^{t-1}) \|/L}.
\end{align}

\noindent
We note that $\mathcal{A}(M)$ represents the foreground area of the given mask $M$, and $\tau$ is a temperature parameter for the Softmax operation.
Note that we also explore alternate methods to compute $\delta^t$ as a function of change in the number of connected components and change in the center of mass of object of interest. Please find details in the Suppl.

%% file: figures/framework.tex
\begin{figure*}[t]
  \centering
  \includegraphics[trim={0 0 0 0}, clip, width=1.\textwidth]{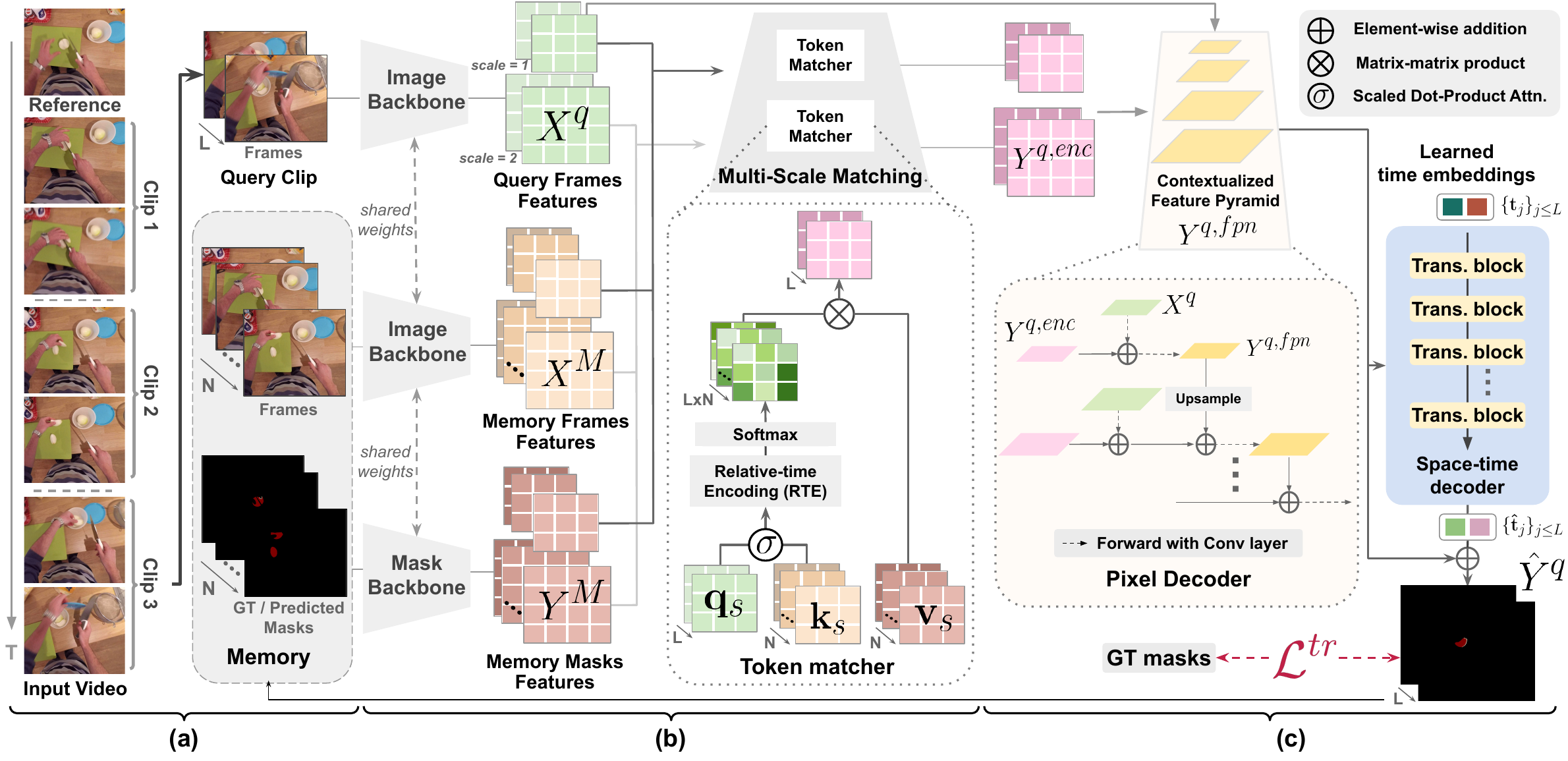}
  \vspace{-6mm}
  \caption{\textbf{Overview of \ourmodel}. We divide an input video into non-overlapping clips of length $L$. For a query clip, we retrieve information on previous clips from our (a) \textit{Clip-based Memory} in the form of frames and predicted (or initial reference) masks. We use a 2D-CNN backbone to obtain features for query frames $X^q$, and features for memory frames and masks, $X^M$ and $Y^M$ respectively. We then use our proposed (b) \textit{Multi-Scale Matching Encoder} to perform dense matching at multiple-scales with frame features of the query clip $X^q$ and memory frames $X^M$, and use the resulting similarity between frames to obtain query clip's mask features $Y^{q, enc}$ as a weighted combination of the memory mask features $Y^M$. In doing so, we modulate the similarity using our proposed multiplicative \textit{Relative-Time Encoding (RTE)} to learn recency of information in memory, thereby facilitating propagation over long-time spans. We then use (c) \textit{Multi-Scale Decoder} to aggregate the resulting query clip's mask features $Y^{q, enc}$  with clip's frame features $X^q$ using Pixel Decoder to give contexualized feature pyramid $Y^{q, fpn}$. Finally, we use Space-Time Decoder to decode mask predictions $\hat{Y}^q$, by refining learned time embeddings on the contextualized feature pyramid $Y^{q, fpn}$. We update the memory with the predictions from last frame (=$L^{th}$ index) in the query clip, implemented as FIFO queue. During training, we use our transformation-aware loss $\mathcal{L}^{tr}$ to form segmentation loss for the entire video.}

  \vspace{-5mm}
  \label{fig:Architecture_1}
\end{figure*}

%% file: sec/4_experiments.tex
\section{Experiments}
We empirically show ability of our method to achieve high-quality video object segmentation and tracking with long-term pixel-level consistency during object transformations. Our approach not only outperforms existing SoTA methods on complex egocentric datasets, VISOR~\cite{darkhalil2022epic} and VOST~\cite{tokmakov2023breaking}, but also achieves competitive results on the generic VOS benchmark, DAVIS'17~\cite{pont20172017}. Lastly, we conduct extensive analyses to validate our design choices.

\subsection{Implementation Details}
We follow a two-stage training protocol similar to prior works~\cite{zhou2022survey, oh2019video, yang2021associating, liang2020video}, where we first pre-train our model using synthetic video sequences generated from static image datasets~\cite{cheng2014global, everingham2010pascal, hariharan2011semantic, shi2015hierarchical, lin2014microsoft}, and then fine-tune on target benchmarks. Consistent with prior works, we use ResNet-50~\cite{he2016deep} as the backbone for image encoder and mask encoder with shared weights. In our Multi-Scale Matching Encoder, if not specified, we use feature maps from last two layers of the visual backbone at strides $32$ and $16$. During training, we use video length of 12, with clip length $L=2$, and a memory bank size of $N=7$. We use  AdamW~\cite{loshchilov2017decoupled} optimizer with $2e^{-5}$ learning rate over 20 epochs with $624$ and $480$ image resolution for VOST and VISOR respectively. For the clip-based memory, 
we retain the ground truth for the initial reference frame and mask as the first element. We perform updates by storing the last frame and its predicted mask features for each clip in the memory in a FIFO fashion. During inference, we process the entire video clip-by-clip to predict the results. Please find more details in Suppl.

\subsection{Datasets and Evaluation}
\vspace{-0.3em}
We evaluate on two benchmark datasets consisting of egocentric videos with pixel-level mask annotations. (1) \textbf{VISOR}~\cite{darkhalil2022epic} contains $7,836$ videos, sourced from Epic-Kitchens~\cite{Damen2022RESCALING} dataset, with an average video length of $12.0$ seconds or $6$ frames per video at $0.5$ FPS,
and (2) \textbf{VOST}~\cite{tokmakov2023breaking} is a recently proposed dataset designed for video object segmentation under complex transformations. The videos in VOST have longer duration and more complex transformations compared to VISOR. VOST contains $713$ videos with an average length of $21.2$ seconds or $106$ frames per video at $5$ FPS. Additionally, we evaluate on a conventional VOS benchmark, DAVIS'17~\cite{pont20172017}.

To evaluate quality of our predicted masks, we use metrics commonly used in Video Object Segmentation, {\em i.e.}, a combination of \textbf{region similarity ($\mathcal{J}$)} and \textbf{contour accuracy ($\mathcal{F}$)}~\cite{tokmakov2023breaking,darkhalil2022epic,perazzi2016benchmark, xu2018youtube, pont20172017}. 
We note that, VOST~\cite{tokmakov2023breaking} introduced an additional metric, $J_{tr}$, designed to evaluate predictions occurring after object transformations. Unlike the standard $J$, $J_{tr}$ evaluates on last $25\%$ of frames in a video sequence, capturing performance of object transformations occurring later in videos. Moreover, following VISOR~\cite{darkhalil2022epic}, we additionally report results on both \textit{seen} and \textit{unseen} subsets of the validation set to assess model's generalization. %

\input{tables/main_visor}
\subsection{Main Results}

\input{figures/demo}

{\bf Comparison to prior works. } 
To validate our ability to track objects in long-videos with complex object transformations, we report results on VOST dataset in Table~\ref{tab:exp:vost}. Our model outperforms the previous best model, AOT~\cite{yang2021associating}, both in terms of $\mathcal{J}$ and $\mathcal{J}_{\text{tr}}$ under the same pre-training and backbone configuration.
Notably, \our achieves comparable performance without relying on the additional DAVIS~\cite{pont20172017} dataset in the pre-training phase.
 
Furthermore, in order to quantitatively investigate various factors affecting the performance, {\em e.g.}, video length (LNG) (\textgreater 20 sec), the presence of multiple instances (MI), and small objects (SM) (\textless 0.5\%\footnote[1]{We follow the definition of small objects provided by the authors of the VOST paper.} frame area), we follow the evaluation protocol outlined in VOST~\cite{tokmakov2023breaking} by evaluating on subsets of validation data representing the above mentioned factors. The comparison is shown in Table~\ref{tab:exp:vost_mode}. Our method achieves best scores compared to all previous methods, reported in VOST~\cite{tokmakov2023breaking}, in terms of the three factors ({\em i.e.}, LNG, MI, and SM), confirming the tracking capability of our model in challenging scenarios. Specifically, we show the highest gain w.r.t.  AOT~\cite{yang2021associating}, in the long video (LNG) case ($+7.2\%$). Our devised loss and multi-scale video transformer not only leverages detailed information necessary for small objects, it reduces tracking drift leading to better and more robust results on long videos.

Additionally, we show comparison on VISOR in Table~\ref{tab:exp:visor}. Our model outperforms the previous best model STM \cite{oh2019video} by $7.1\%$ on $\mathcal{J}\&\mathcal{F}$ and $10.3\%$ on $\mathcal{J}\&\mathcal{F}_{\text{unseen}}$.
We note that, in \textit{unseen} setting, STM \cite{oh2019video} reports a considerable $4.6\%$ ($75.8 \rightarrow$ 71.2) drop in $\mathcal{J}\&\mathcal{F}_{\text{unseen}}$, whereas our method only shows $1.4\%$ ($82.9 \rightarrow$ 81.5) drop. This suggests stronger generalization ability and robustness of our proposed approach.

\vspace{0.05in}
\noindent
{\bf Qualitative results.}
We compare our proposed method with the prior best model, AOT, on VOST. Qualitative examples are shown in Fig.~\ref{fig:demo}. We select an example from the validation set, present in both the long video and multi-instance subset. We observe that AOT's outputs struggle to maintain long-term consistency for the same object across the long sequence, resulting in confusion between different parts of multiple ``onions" while cutting one. In contrast, our method, which incorporates a clip-based memory module and the multi-scale matching and decoding strategy, effectively preserves consistency across long sequences for the objects. This enables us to generate accurate masks for each object without mixing them.
For video version of this example and additional qualitative results, please refer to our supplementary materials.

\input{tables/ablations/memory}

\subsection{Ablation Study and Model Analysis}
\label{sec:exps_analysis}
To verify the effectiveness of key design elements in our model, we conduct ablation studies and model analysis with different settings on VOST~\cite{tokmakov2023breaking}. The results are shown in Table~\ref{tab:ablation:memory}. We consider six different cases,

\input{figures/vis_RTE_short}
\vspace{0.1in}
\noindent
\textbf{Relative-time encoding (RTE).} Rows ($1$-$3$) show that our proposed time-coded memory improves the results. Moreover, we show ablation w.r.t. conventional additive relative positional embeddings~\cite{raffel2020exploring} (row $2$), where ours modulates the learning in a multiplicative manner (row $3$) while allowing for different clip lengths. It shows on-par performance in the mIoU ($\mathcal{J}$), while in the challenging scenarios, $\mathcal{J}_{tr}$, our encoding improves by up to $1.7\%$, verifying its effectiveness. This is likely attributed to the stronger impact of multiplicative modulation vs. additive one which was also observed in other domains in convolutional architectures~\cite{feichtenhofer2017spatiotemporal}. We visualize the learned embeddings ($e_n \in \mathbb{R}^{1 \times n}$) of our RTE in \cref{exp:vis_RTE_short} in the case of $n = \{4,5,6\}$, where $n$ is the number of frames in memory. The results indicate that our RTE equips the model with the flexibility to determine the importance of each frame in memory during the matching process. Notably, we observe that a query frame relies more on the two closest frames (= indices $0$ and $1$) compared to those farther away. Moreover, regardless of the number of frames in the memory, the first frame (= last index) always holds significance as it represents the ground truth / reference mask. More results in Suppl.

\vspace{0.1in}
\noindent
\textbf{Transformation-aware reweighting.} The rows ($4$ vs. $5$) and rows ($3$ vs. $6$) show that the transformation-aware loss further improve the performance on $\mathcal{J}_{tr}$ by $\sim 2\%$, confirming the efficacy of our proposed loss which incorporates a ``soft" hard example mining on video-level loss during training.

\vspace{0.1in}
\noindent
\textbf{Multi-scale matching encoder-decoder.} Rows ($7$-$8$) show that simply using a single-scale encoder-decoder setup (=$1$), where both the encoder matching and decoding is performed at single-scale
results in significantly lower performance, confirming the effectiveness of our multi-scale matching and decoding framework. Training on three-scales (row $10$) faces GPU memory limitations which requires lowering video length and memory bank size to make fair comparisons. Overall, we found two scales to perform better (rows $9$-$10$), likely due to lower discriminability at higher scales beyond two, resulting in inefficient matching. Thus, we choose two scales which gives the best trade-off in terms of accuracy and memory efficiency. Although, prior works~\cite{seong2021hierarchical,yang2021collaborative} have explored multi-scale attention matching beyond two-scales, we note that our paper studies multiple-scales at both the encoder and decoder level together in a unified framework, and any efficient attention mechanisms can be readily adapted.  We also qualitatively explore the effectiveness of this multi-scale design in \cref{fig:demo_attn}. We select a video featuring small objects, and plot attention heat maps produced by our multi-scale matching encoder. We observe that our model is able to finely match small objects in feature maps at multiple-scales. In particular, in the third column (or frame), we observe that the attention at the coarse-scale (=$1$) is scattered on human hand and other parts of the frame, however on finer-scale (=$2$), it successfully attends to the object of interest. These findings illustrate the ability of our multi-scale matching to capture small objects,
additional results is in the Suppl.

\vspace{0.1in}
\noindent
\textbf{Video length.} We examine the impact of video length used during training. Rows ($11$-$14$) show that a video length of $12$ achieves the best performance when compared to shorter ({\em i.e.}, $9$) and longer ({\em i.e.}, $15$) video lengths.

\vspace{0.1in}
\noindent
\textbf{Bank size ($N$).} Rows ($15$-$18$) indicate that increase in the memory bank size results in better performance. However, this expansion requires extending the video length to store more frames in memory, demanding additional computational resources. GPU memory limitations cap our memory bank size at $7$. We believe that training on GPUs that can accommodate larger memory bank, could lead to further improvements. 

\vspace{0.1in}
\noindent
\textbf{Clip length ($L$).} Rows ($19$-$23$) show that clip-level inference and memory updates perform better than frame-level ($L\!=\!1$). In addition, PCVOS~\cite{park2022per} found that using even longer clip lengths ($>10$) results in decrease in performance. In our case, building on the above trade-off on clip length, and taking into account our design choices (multi-scale matching and decoding) together with limitations on GPU memory, we use clip length of $2$ for our final model.

Finally, based on the above observations, we choose the final model configuration for optimal performance during training: a video length of $12$ frames, a clip length of $2$, a memory bank size of $7$, with RTE, with multi-scale matching using $2$ scales and with transformation-aware loss. We provide further ablations on the relative-time encoding and transformation-aware loss in the Suppl. materials.

\input{tables/main_davis}

\vspace{-4mm}
\subsection{Performance Comparison on Conventional VOS Datasets}

We also verify the effectiveness of our model on DAVIS'17~\cite{pont20172017} benchmark.
Following standard training protocol~\cite{seong2021hierarchical,yang2021collaborative}, we pre-train on Static datasets~\cite{cheng2014global, everingham2010pascal, hariharan2011semantic, shi2015hierarchical, lin2014microsoft}. First, we report fine-tuning performance on DAVIS'17 only in Table~\ref{tab:exp:davis}. Our approach achieves best results on overall scores $\mathcal{J}\&\mathcal{F}$ and 1.7\% improvement on the $\mathcal{F}$ metric. Second, we draw comparisons with HMMN~\cite{seong2021hierarchical} on small objects, where the latter make use of efficient multi-scale memory matching. We divide DAVIS'17 into subsets based on object size (similar to the VOST dataset). Table~\ref{tab:exp:davis_subset} shows that our model is better equipped to capture small-sized objects on conventional benchmark, demonstrating the efficacy of our unified multi-scale encoder-decoder framework. Lastly, we provide the results of our model when trained on both YouTubeVOS and DAVIS'17 jointly and evaluated on the DAVIS'17 val set. Table~\ref{tab:exp:davis_youtube} show that our model is close to or even outperform previous works, verifying the generalization of our approach. Note that in the challenging VOST our method outperforms XMem with a considerable margin.

%% file: tables/main_visor.tex
\begin{table}[t]
    \caption{\textbf{Performance comparison on complex egocentric benchmarks.} (a) \textbf{Val performance on VOST}. Our method outperforms the best reported results on both $\mathcal{J}$ and $\mathcal{J}_{\text{tr}}$ metrics. We note that Static represents a collection of image datasets~\cite{cheng2014global, everingham2010pascal, hariharan2011semantic, shi2015hierarchical, lin2014microsoft}. Performance numbers on the prior works are reported from VOST. $^\dag$ means results reproduced by us using official code. (b) \textbf{Quantitative analysis of various factors on VOST val set.} We report $\mathcal{J}_{\text{tr}}$ score for each case. LNG: the subset comprising long videos, MI: the subset with multi-instances and SM: the subset with small objects. We report the gain, (.), w.r.t the best reported model, AOT~\cite{yang2021associating}. $^\dag$ denotes results reproduced by us using official code. (c) \textbf{Val performance on VISOR}. Under similar pre-training and backbone configurations, our method outperforms best reported results, with a marginal drop in \textit{unseen} performance (=$1.4\%$) compared to STM~\cite{oh2019video,darkhalil2022epic} (=$4.6\%$). Performance numbers on STM~\cite{oh2019video} are reported from VISOR~\cite{darkhalil2022epic}.}
    \vspace{-5mm}
    \begin{subtable}[t]{.5\linewidth}
    \centering
    \resizebox{0.95\columnwidth}{!}{
    \begin{tabular}[t]{l|c|cc}
    \toprule
    \multicolumn{1}{c|}{\multirow{2}{*}{Approach}} & \multirow{2}{*}{Pre-training} & \multicolumn{2}{c}{VOST} \\ \cline{3-4} 
    \multicolumn{1}{c|}{} &  & $\mathcal{J}_{\text{tr}}$ & $\mathcal{J}$  \\ \hline
    OSMN-Match~\cite{yang2018efficient} & Static + DAVIS~\cite{pont20172017} & 7.0 & 8.7  \\
    OSMN-Tune~\cite{yang2018efficient} & Static + DAVIS & 17.6 & 23.0  \\
    CRW~\cite{jabri2020space} & IN1K~\cite{deng2009imagenet} + DAVIS & 13.9 & 23.7  \\
    HODOR-Img~\cite{athar2022hodor} & COCO~\cite{lin2014microsoft} + DAVIS & 13.9 & 26.2 \\
    HODOR-Vid~\cite{athar2022hodor} & COCO + DAVIS & 25.4 & 37.1 \\
    CFBI~\cite{yang2020collaborative} & IN1K + COCO + DAVIS & 32.0 & 45.0 \\
    CFBI+~\cite{yang2021collaborative} & Static + DAVIS & 32.6 & 46.0  \\
    XMem~\cite{cheng2022xmem} & Static + DAVIS & 33.8 & 44.1  \\
    AOT$^\dag$~\cite{yang2021associating} & Static & 35.1 & 47.1 \\
    AOT~\cite{yang2021associating} & Static + DAVIS & 36.4 & 48.7 \\ 
    \midrule
    \ourmodel (Ours) & Static & 36.5 & 48.2  \\
    \ourmodel (Ours) & Static + DAVIS & \textbf{37.7} & \textbf{49.3}  \\
    \bottomrule
    \end{tabular}
    }
    \caption{\textbf{Val performance on VOST}}
    \label{tab:exp:vost}
    \end{subtable}
    \begin{subtable}{.5\linewidth}
        \begin{subtable}[t]{1.0\linewidth}
        \centering
        \resizebox{0.95\columnwidth}{!}{
        \begin{tabular}[t]{l|cccc|c}
        \toprule
        \multicolumn{1}{c|}{} & \begin{tabular}[c]{@{}c@{}}OSMN\\ Tune~\cite{yang2018efficient}\end{tabular} & CFBI+~\cite{yang2021collaborative} & \begin{tabular}[c]{@{}c@{}}HODOR\\ Vid~\cite{athar2022hodor}\end{tabular} & AOT~\cite{yang2021associating} & \begin{tabular}[c]{@{}c@{}}\ourmodel (Ours)\\ (\textcolor{blue}{diff with AOT~\cite{yang2021associating}})\end{tabular} \\ \hline
        All & 17.6 & 32.6 & 25.4 & 36.4 & \textbf{37.7} \textcolor{blue}{(+1.3)} \\ \rowcolor[HTML]{ECF4FF} 
        LNG & 12.4 & 30.4 & 25.0 &  34.7 & \textbf{41.9} \textcolor{blue}{(+7.2)} \\
        MI & 14.7 & 26.4 & 20.6 & 27.2 & \textbf{29.2} \textcolor{blue}{(+2.0)} \\ \rowcolor[HTML]{ECF4FF} 
        SM & 14.4 & 23.3 & 16.6 & 24.7 & \textbf{28.4} \textcolor{blue}{(+3.7)} \\
        \bottomrule
        \end{tabular}
        }
        \caption{\textbf{Quantitative analysis of factors on VOST.}}
        \label{tab:exp:vost_mode}
        \end{subtable}
        \begin{subtable}{1.0\linewidth}
        \centering
        \resizebox{0.95\columnwidth}{!}{
        \begin{tabular}{lcccccc}
        \toprule
        \multicolumn{1}{c|}{Approach} & \begin{tabular}[c]{@{}c@{}}Pre-train\\ Static\end{tabular} & \multicolumn{1}{c|}{\begin{tabular}[c]{@{}c@{}}Fine-Tune\\ VISOR\end{tabular}} & $\mathcal{J}\&\mathcal{F}$ & $\mathcal{J}$ & $\mathcal{F}$ & $\mathcal{J}\&\mathcal{F}_{\text{unseen}}$ \\ \hline
        \multicolumn{1}{l|}{STM \cite{oh2019video, darkhalil2022epic}} & \checkmark & \multicolumn{1}{c|}{} & 56.9 & 55.5 & 58.2 & 48.1 \\
        \multicolumn{1}{l|}{STM \cite{oh2019video, darkhalil2022epic}} & \checkmark & \multicolumn{1}{c|}{\checkmark} & 75.8 & 73.6 & 78.0 & 71.2 \\ 
        \midrule
        \multicolumn{1}{l|}{\ourmodel (Ours)} & \checkmark & \multicolumn{1}{c|}{} & 63.7 & 61.5 & 65.9 & 56.3 \\
        \multicolumn{1}{l|}{\ourmodel (Ours)} & \checkmark & \multicolumn{1}{c|}{\checkmark} & \textbf{82.9} & \textbf{80.6} & \textbf{85.1} & \textbf{81.5} \\
        \bottomrule
        \end{tabular}%
        }
        \caption{\textbf{Val performance on VISOR}}
        \label{tab:exp:visor}
        \end{subtable}
    \end{subtable}
    \vspace{-3em}
\end{table}

%% file: figures/demo.tex
\begin{figure}[t]
  \centering
  \includegraphics[page=1,trim={5 590 40 0}, clip, width=1.\columnwidth]{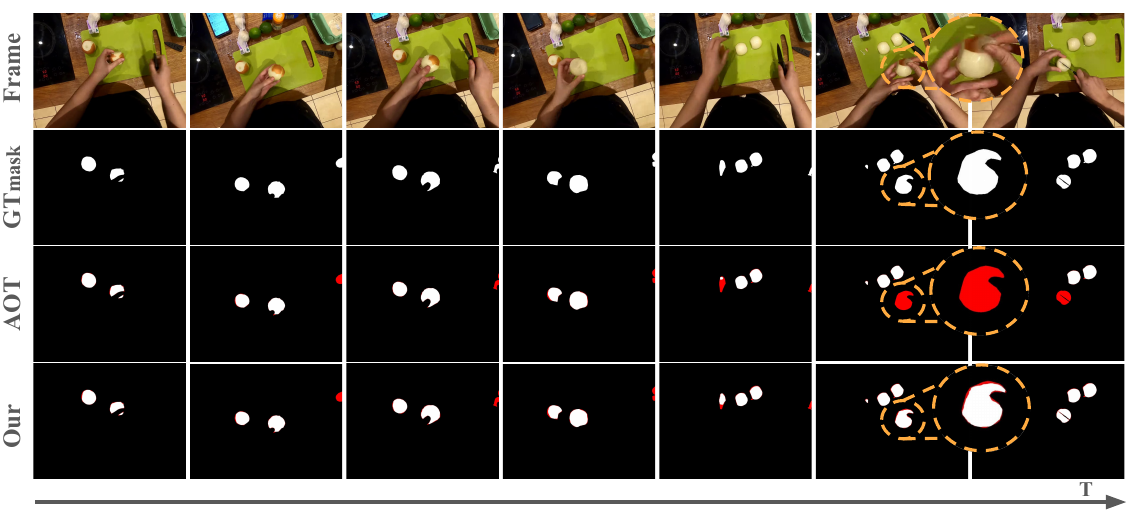}
  \vspace{-7mm}
  \caption{{\bf Qualitative comparison on VOST.} Best viewed in color; red indicates incorrect predictions. Our method is able to track and delineate the object's boundary with fine details under complex deformation, e.g. cutting onions, compared to best prior work, AOT~\cite{yang2021associating}.}
  \label{fig:demo}
  \vspace{-6mm}
\end{figure}

%% file: tables/ablations/memory.tex
\begin{table}[t]
    \caption{\textbf{Ablation study on VOST val set}. We examine various parameter combinations, where $N$ represents the bank size of the memory, $L$ is the clip length, RTE denotes multiplicative relative time encoding and Add. denotes additive relative positional encoding. Note that numbers with $\underline{\text{underline}}$ means best performance within ablation group, and \textbf{bold} the overall best.}
    \vspace{-4mm}

    \begin{subtable}[t]{.495\linewidth}
    \centering
    \resizebox{0.96\columnwidth}{!}{%
    \begin{tabular}[t]{c|c|c|c|c|c|c|cc}
    \toprule
    \multicolumn{1}{c|}{\multirow{2}{*}{\#}} & \multicolumn{1}{c|}{\multirow{1}{*}{Video}} & \multicolumn{1}{c|}{\multirow{2}{*}{$L$}} & \multicolumn{2}{c|}{Memory} & $\mathcal{L}^{\text{tr}}_{\text{focal}}$ & Scales & \multirow{2}{*}{$\mathcal{J}_{\text{tr}}$} & \multirow{2}{*}{$\mathcal{J}$} \\ \cline{4-5}
    \multicolumn{1}{c|}{} & \multicolumn{1}{c|}{length} & \multicolumn{1}{c|}{} & \multicolumn{1}{c|}{$N$} & \multicolumn{1}{c|}{RTE} & & &  & \\
    
    \hline
    \multicolumn{9}{c}{w/ or w/o RTE} \\ \hline
    \multicolumn{1}{c|}{(1)} & \multicolumn{1}{c|}{12} & \multicolumn{1}{c|}{2} & 7 & \multicolumn{1}{c|}{} & & 2 &  31.9 & 45.5\\ %
    \multicolumn{1}{c|}{(2)} & \multicolumn{1}{c|}{12} & \multicolumn{1}{c|}{2} & 7 & \multicolumn{1}{c|}{Add.} & & 2 & 32.0 & \underline{47.4} \\
    \multicolumn{1}{c|}{(3)} & \multicolumn{1}{c|}{12} & \multicolumn{1}{c|}{2} & 7 & \multicolumn{1}{c|}{$\checkmark$} & & 2 & \underline{33.7} & \underline{47.2} \\
    
    \hline
    \multicolumn{9}{c}{w/ or w/o Transformation-aware Loss $\mathcal{L}^{\text{tr}}_{\text{focal}}$} \\ \hline
    \multicolumn{1}{c|}{(4)} & \multicolumn{1}{c|}{12} & \multicolumn{1}{c|}{2} & 7 & \multicolumn{1}{c|}{{}} &  & 2 & 31.9 & 45.5 \\ %
    \multicolumn{1}{c|}{(5)} & \multicolumn{1}{c|}{12} & \multicolumn{1}{c|}{2} & 7 & \multicolumn{1}{c|}{} & {$\checkmark$} & 2 & 33.5 & 46.6 \\ 
    \multicolumn{1}{c|}{(6)} & \multicolumn{1}{c|}{12} & \multicolumn{1}{c|}{2} & 7 & \multicolumn{1}{c|}{$\checkmark$} & {$\checkmark$} & 2 & \textbf{36.5} & \textbf{48.2} \\
    
    \hline
    \multicolumn{9}{c}{w/ or w/o multi-scale matching and decoding} \\ \hline
    \multicolumn{1}{c|}{(7)} & \multicolumn{1}{c|}{12} & \multicolumn{1}{c|}{2} & 7 &  & \multicolumn{1}{c|}{} & 1 (w/o) & 22.5 & 33.9  \\ 
    \multicolumn{1}{c|}{(8)} & \multicolumn{1}{c|}{12} & \multicolumn{1}{c|}{2} & 7 & \multicolumn{1}{c|}{} & & 2 (w/) & \underline{31.9} & \underline{45.5} \\
    
    \hline
    \multicolumn{9}{c}{Number of scales used in multi-scale matching} \\ \hline
    \multicolumn{1}{c|}{(9)} & \multicolumn{1}{c|}{6} & \multicolumn{1}{c|}{2} & 4 &  & \multicolumn{1}{c|}{} & 2 & 29.2 & 43.9 \\
    \multicolumn{1}{c|}{(10)} & \multicolumn{1}{c|}{6} & \multicolumn{1}{c|}{2} & 4 &  & \multicolumn{1}{c|}{} & 3 & 27.2 & 40.2 \\
    
    \bottomrule
    \end{tabular}
    }
    \end{subtable} 
    \begin{subtable}[t]{.505\linewidth}
    \centering
    \resizebox{0.94\columnwidth}{!}{%
    \begin{tabular}[t]{c|c|c|c|c|c|c|cc}
    \toprule
    \multicolumn{1}{c|}{\multirow{2}{*}{\#}} & \multicolumn{1}{c|}{\multirow{1}{*}{Video}} & \multicolumn{1}{c|}{\multirow{2}{*}{$L$}} & \multicolumn{2}{c|}{Memory} & $\mathcal{L}^{\text{tr}}_{\text{focal}}$ & Scales & \multirow{2}{*}{$\mathcal{J}_{\text{tr}}$} & \multirow{2}{*}{$\mathcal{J}$} \\ \cline{4-5}
    \multicolumn{1}{c|}{} & \multicolumn{1}{c|}{length} & \multicolumn{1}{c|}{} & \multicolumn{1}{c|}{$N$} & \multicolumn{1}{c|}{RTE} & & &  & \\
    
    \hline
    \multicolumn{9}{c}{w/ different video lengths} \\ \hline
    \multicolumn{1}{c|}{(11)} & \multicolumn{1}{c|}{9} & \multicolumn{1}{c|}{2} & 3 & \multicolumn{1}{c|}{$\checkmark$} & & 2 & 30.7 & 44.5 \\
    \multicolumn{1}{c|}{(12)} & \multicolumn{1}{c|}{12} & \multicolumn{1}{c|}{2} & 3 & \multicolumn{1}{c|}{$\checkmark$} & & 2 & \underline{32.9} & \underline{45.4} \\ \midrule
    \multicolumn{1}{c|}{(13)} & \multicolumn{1}{c|}{12} & \multicolumn{1}{c|}{2} & 7 & \multicolumn{1}{c|}{$\checkmark$} & & 2 & \underline{33.7} & \underline{47.2} \\
    \multicolumn{1}{c|}{(14)} & \multicolumn{1}{c|}{15} & \multicolumn{1}{c|}{2} & 7 & \multicolumn{1}{c|}{$\checkmark$} & & 2 & 32.6 & 47.0 \\
    
    \hline
    \multicolumn{9}{c}{w/ different memory bank sizes, $N$} \\ \hline
    \multicolumn{1}{c|}{(15)} & \multicolumn{1}{c|}{12} & \multicolumn{1}{c|}{2} & 1 & \multicolumn{1}{c|}{$\checkmark$} & & 2 & 27.4 & 40.5 \\
    \multicolumn{1}{c|}{(16)} & \multicolumn{1}{c|}{12} & \multicolumn{1}{c|}{2} & 3 & \multicolumn{1}{c|}{$\checkmark$} & & 2 & 32.9 & 45.4 \\
    \multicolumn{1}{c|}{(17)} & \multicolumn{1}{c|}{12} & \multicolumn{1}{c|}{2} & 5 & \multicolumn{1}{c|}{$\checkmark$} & & 2 & 32.0 & 45.8 \\
    \multicolumn{1}{c|}{(18)} & \multicolumn{1}{c|}{12} & \multicolumn{1}{c|}{2} & 7 & \multicolumn{1}{c|}{$\checkmark$} & & 2 & \underline{33.7} & \underline{47.2} \\
    
    \hline
    \multicolumn{9}{c}{w/ different clip lengths, $L$} \\ \hline
    \multicolumn{1}{c|}{(19)} & \multicolumn{1}{c|}{12} & \multicolumn{1}{c|}{1} & 7 & \multicolumn{1}{c|}{$\checkmark$} & & 2 & 32.3 & 47.1 \\
    \multicolumn{1}{c|}{(20)} & \multicolumn{1}{c|}{12} & \multicolumn{1}{c|}{2} & 7 & \multicolumn{1}{c|}{$\checkmark$} & & 2 & \underline{33.7} & \underline{47.2} \\ \midrule
    \multicolumn{1}{c|}{(21)} & \multicolumn{1}{c|}{12} & \multicolumn{1}{c|}{2} & 5 & \multicolumn{1}{c|}{$\checkmark$} & & 2 & 32.0 & 45.8 \\
    \multicolumn{1}{c|}{(22)} & \multicolumn{1}{c|}{12} & \multicolumn{1}{c|}{3} & 5 & \multicolumn{1}{c|}{$\checkmark$} & & 2 & \underline{34.9} & \underline{46.5} \\ 
    \multicolumn{1}{c|}{(23)} & \multicolumn{1}{c|}{16} & \multicolumn{1}{c|}{4} & 5 & \multicolumn{1}{c|}{$\checkmark$} & & 2 & 31.3 & 44.4  \\ 
    \bottomrule
    \end{tabular}
    }
    \end{subtable} 
    \label{tab:ablation:memory}
    \vspace{-5mm}
\end{table}

%% file: figures/vis_RTE_short.tex
\begin{figure*}[t]
  \begin{subfigure}[t]{0.5\textwidth}
  \centering
  \includegraphics[page=1,trim={5 10 30 0}, clip, width=0.95\columnwidth]{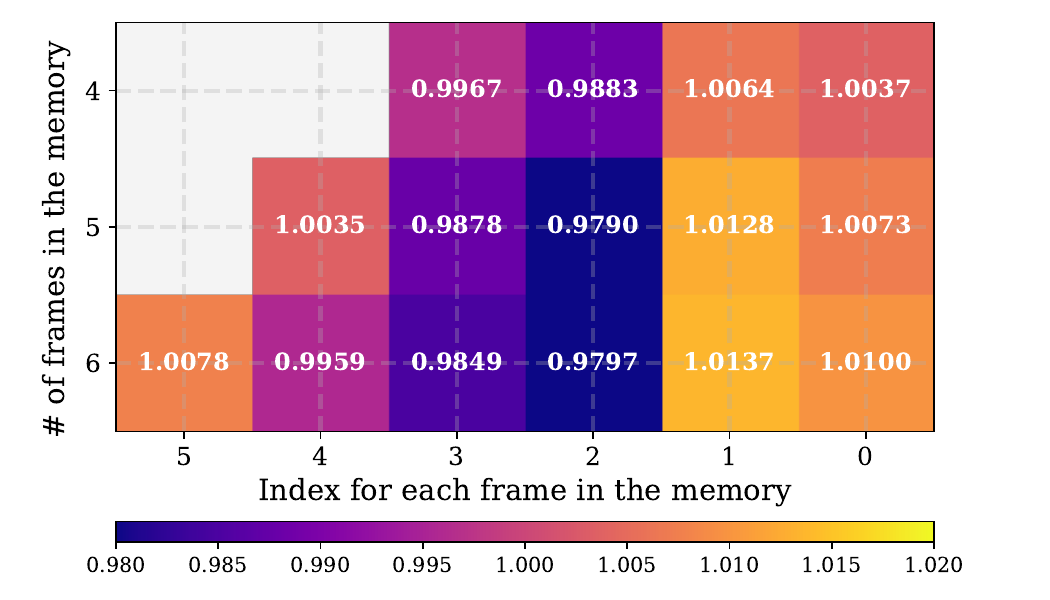}
  \caption{ }
  \label{exp:vis_RTE_short}
  \end{subfigure}%
  \begin{subfigure}[t]{0.5\textwidth}
  \centering
  \includegraphics[page=2,trim={0 670 290 0}, clip, width=\columnwidth]{figures/sources/demo.pdf}
  \caption{ }
  \label{fig:demo_attn}
  \end{subfigure}
  \vspace{-3mm}
  \caption{(a) \textbf{Visualization of RTE}. Value in each grid represents the learned importance score for each frame w.r.t. different number of frames in the memory ($n$). Lighter colors indicate higher scores. In each row, index $0$ denotes the nearest frame, last index denotes the first frame, and remaining indices denote intermediate frames in memory. (b) \textbf{Visualization of the attention maps} in the multi-scale matching modules. Scale 1 has a resolution of $\frac{1}{32}$ of the input frame, and scale 2 is at $\frac{1}{16}$ resolution. Best viewed in color; lighter colors indicate higher attention scores. The red box in the first frame denotes the object of interest in the query frame.}
  \vspace{-7mm}
\end{figure*}

%% file: tables/main_davis.tex
\begin{table}[t]
    \caption{\textbf{Performance comparison on conventional datasets}. (a) DAVIS'17 val. set. (b) DAVIS'17 val. subsets w.r.t. object sizes. (c) DAVIS'17 val. set (jointly trained with YouTubeVOS dataset). Best results highlighted in \textbf{Bold} and $\underline{\text{underline}}$ respectively.}
    \vspace{-4mm}
    \centering
    \begin{subtable}[t]{.344\linewidth}
        \begin{subtable}[t]{0.97\linewidth}
        \centering
        \resizebox{0.95\columnwidth}{!}{
        \begin{tabular}[t]{l|c|c|c} 
        \toprule
        \multicolumn{1}{c|}{Approach} & $\mathcal{J}\&\mathcal{F}$ & $\mathcal{J}$ & $\mathcal{F}$ \\ \hline
        SSTVOS~\cite{duke2021sstvos} & 78.4 & 75.4 & 81.4 \\
        AOT-S~\cite{yang2021associating} & 79.2 & 76.4 & 82.0 \\
        TBD~\cite{cho2022tackling} & 80.0 & 77.6 & 82.3 \\
        HMMN~\cite{seong2021hierarchical} & 80.4 & \textbf{77.7} & 83.1 \\ 
        \midrule
        \ourmodel (Ours) & \textbf{81.0} & 77.1 & \textbf{84.9} \\
        \bottomrule
        \end{tabular}%
        }
        \caption{Pre-trained on Static and fine-tuned on DAVIS'17 only.}
        \label{tab:exp:davis}
        \end{subtable}
        \begin{subtable}{1.0\linewidth}
        \centering
        \resizebox{0.95\columnwidth}{!}{
        \begin{tabular}{c|c|c|cc}
        \toprule
        \multicolumn{1}{c|}{\multirow{2}{*}{Approach}} & \multicolumn{3}{c}{Object size} \\ \cline{2-4} 
        & $<1\%$ & $<0.5\%$ & $<0.3\%$ \\ \midrule
        HMMN~\cite{seong2021hierarchical} & 65.1 & 52.3 & 17.4 \\
        \ourmodel (Ours) & \textbf{69.5} & \textbf{63.0} & \textbf{42.6} \\
        \bottomrule
        \end{tabular}
        }
        \caption{Performance breakdown on DAVIS'17 val set w.r.t. small object sizes.}
        \label{tab:exp:davis_subset}
        \end{subtable}
    \end{subtable}
    \begin{subtable}[t]{.41\linewidth}
    \centering
    \resizebox{0.95\columnwidth}{!}{
    \begin{tabular}[t]{l|c|c|c}
    \toprule
    \multicolumn{1}{c|}{Approach} & $\mathcal{J}\&\mathcal{F}$ & $\mathcal{J}$ & $\mathcal{F}$ \\ \hline
    STM~\cite{oh2019video, darkhalil2022epic} & 81.8 & 79.2 & 84.3 \\
    CFBI~\cite{yang2020collaborative} & 81.9 & 79.1 & 84.6 \\
    CFBI+~\cite{yang2020collaborative} & 82.9 & 80.1 & 85.7 \\
    STCN~\cite{cheng2021rethinking} & \underline{85.4} & 82.2 & \underline{88.6} \\
    HMMN~\cite{seong2021hierarchical} & 84.7 & 81.9 & 87.5 \\
    AOT~\cite{yang2021associating} & 84.9 & 82.3 & 87.5 \\
    RDE~\cite{li2022recurrent} & 84.2 & 80.8 & 87.5 \\
    XMem~\cite{cheng2022xmem} & \textbf{86.2} & \textbf{82.9} & \textbf{89.5} \\
    DeAOT~\cite{yang2022decoupling} & 85.2 & 82.2 & 88.2 \\ 
    \midrule
    \ourmodel (Ours) & \underline{85.3} & \underline{82.5} & 88.2 \\
    \bottomrule
    \end{tabular}
    }
    \caption{Pre-trained on Static and jointly fine-tuned on YouTubeVOS and DAVIS'17.}
    \label{tab:exp:davis_youtube}
    \end{subtable}
    \label{tab:exp:davis_all}
    \vspace{-8mm}
\end{table}

%% file: sec/5_conclusion.tex
\section{Conclusion}
\vspace{-0.05in}
In this paper, we proposed a novel multi-scale memory matching and decoding scheme with clip-based time-coded memory for Semi-VOS task, designed to cope with complex object transformations, small objects and long video tracking. Moreover, we proposed a technique for hard examples mining through a novel transformation-aware loss. Our results show that the proposed model outperforms previous state-of-the-art on Semi-VOS across two benchmarks, with significant gains on VISOR benchmark up to 10\%. %

%% file: sec/suppl.tex
\clearpage
\setcounter{page}{1}
\def\thesection{\Alph{section}}
\renewcommand{\thetable}{A\arabic{table}}
\renewcommand{\thefigure}{A\arabic{figure}}
\title{TAM-VT: Transformation-Aware Multi-scale Video Transformer for Segmentation and Tracking \\ (supplementary material)} 

\titlerunning{TAM-VT: Transformation-Aware Multi-scale Semi-VOS}

\author{
Raghav Goyal$^{*1,2}$ \hspace{0.25in} Wan-Cyuan Fan$^{*1,2}$ \hspace{0.25in} Mennatullah Siam$^{4}$ \\ Leonid Sigal$^{1,2,3}$ 
}
\authorrunning{Goyal and Fan et al.}

\institute{$^1$University of British Columbia \hspace{0.2in}
$^2$Vector Institute for AI \hspace{0.2in}
$^3$CIFAR AI Chair\\
$^4$Ontario Tech University
}

\maketitle

\begin{abstract}
This document provides additional material that is supplemental to our main submission. Section~\ref{sec:video} describes the associated supplemental video. Section~\ref{sec:imp} includes additional implementation and dataset details, followed by Section~\ref{sec:exps} for additional experimental results and ablation studies. Finally, Section~\ref{sec:societal} details the societal impact of our work as standard practice in computer vision research.
\end{abstract}

\section{Supplemental Video}
\label{sec:video}
We include an accompanying supplemental video - {\tt video\_demo.mp4} - as part of the supplemental materials. In this video we show qualitative segmentation results of our approach on the two benchmarks VOST~\cite{tokmakov2023breaking} and VISOR~\cite{darkhalil2022epic}. For VOST, we show predictions on three videos showing our method's capability on small and multiple instances of objects compared to the previous best state-of-the-art method, AOT \cite{yang2021associating}. 
We additionally include a failure case. For VISOR, we present three videos demonstrating our method's tracking capabilities compared to the previous best method, STM~\cite{oh2019video}, and include a failure case as well. Furthermore, we evaluate our model on generic videos sampled from YouTube, with the results presented in the final section of the video. The video is in MP4 format and is $4$ minutes $17$ seconds long. The codec used for creating the provided video is H.264 (x264). Please note that due to file size limitations for supplementary material, we provide the low-resolution version, while a higher-resolution demo video will be made publicly available upon acceptance.

\section{Implementation and dataset details}
\label{sec:imp}

\paragraph{Training.} During training, we use AdamW~\cite{loshchilov2017decoupled} optimizer with weight decay $1e^{-4}$, learning rate $1e^{-5}$ for the backbone and $2e^{-4}$ for the rest of the model. We train for 20 epochs and reduce learning rate by $1/10^{th}$ factor at the $8^{th}$ and $16^{th}$ epoch.
Similar to AOT~\cite{yang2021associating}, we use random resizing and cropping for data augmentation, avoiding additional augmentations to ensure a fair comparison. We uniformly use batch size of $1$ per GPU, and conduct experiments on VOST and VISOR datasets using $4$ Nvidia A40 and $8$ Nvidia T4 GPUs respectively. We use video lengths of $12$ and $6$ for VOST and VISOR respectively, necessitating the use of higher GPU memory in the case of VOST compared to VISOR.
The input image resolution is set to $480$ and $624$ pixels on shorter side (keeping the aspect ratio same) on VISOR and VOST respectively.

\paragraph{Architecture.}
Our model is built on ResNet-50~\cite{he2016deep} backbone, which excludes dilation in the last convolutional block compared to STM and related approaches~\cite{oh2019video,park2022per}. The backbone is shared for encoding both the images and masks. Keeping in line with TubeDETR~\cite{yang2022tubedetr}, we use fixed-sinusoidal 2D positional embedding for both image and mask features. For multi-scale query matching, we use a multi-head attention layer~\cite{kim2023universal} with a hidden size of $256$ and $4$ heads for each scale. We perform LayerNorm~\cite{ba2016layer} on queries and keys before multi-head attention, and on output after the multi-head attention. We use GELU~\cite{hendrycks2016gaussian} activation function on the residual connection for the output, and use a $10\%$ attention dropout to reduce overfitting. 

Our decoder consists of a pixel decoder~\cite{cheng2022masked} and a space-time transformer decoder~\cite{yang2022tubedetr} used after the multi-scale matching encoder. The pixel decoder~\cite{cheng2022masked} produces feature-pyramid at four scales using a combination of lateral and output {\tt Conv} layers, with the hidden dimension / output channel size of $256$.
And the space-time transformer decoder~\cite{yang2022tubedetr} incorporates $6$ layers of multi-head attention, comprising $8$ heads with a hidden dimension of $256$. For the output, we use {\tt ReLU} as the activation function and maintain a $10\%$ dropout rate during training.

\input{tables/suppl_dataset}

\paragraph{Dataset.}
The data statistics for VOST~\cite{tokmakov2023breaking} and VISOR~\cite{darkhalil2022epic} are outlined in~\cref{suppl:tab:datasets}. Notably, the videos in VOST are twice as long in duration compared to VISOR. In particular, for VOST, the average length of videos is $21.2$ seconds, annotated at $5$ frames-per-second (fps). This amounts to approximately $106$ annotated frames per video on average. In comparison, VISOR includes $6$ annotated frames per video with a much lower $0.5$ annotation frame rate (fps).

Both datasets focus on egocentric videos, thereby resulting in relatively smaller average object sizes in comparison to the conventional VOS datasets~\cite{pont20172017}. Specifically, VOST's average object size equates to $2.57\%$ of the frame size, whereas VISOR's average object size corresponds to $6.67\%$ of the frames. This indicates that designing a model tailored for objects of various sizes (including small objects) is essential in egocentric videos, which is tied to various applications in robotics and augmented/virtual reality.

\clearpage
\section{Additional experimental results}
\label{sec:exps}

In this section, we provide additional quantitative and qualitative results, along with additional visualization and analysis.

\subsection{Additional quantitative results}
\paragraph{VOST subsets ablations.}
\input{figures/suppl_subset_objectsize}
In the main paper, we evaluate across various subsets of validation data to measure capability of the model across different aspects of the problem, which includes long videos (LNG) and small objects (SM) subsets. 
To further demonstrate our model's performance in these challenging scenarios, we provide a breakdown of results on the subsets at a granular level. The results based on video length subsets is shown in~\cref{suppl:exp:subset_video}, illustrating the $\mathcal{J}_{\text{tr}}$ scores on different video length ranges. Notably, while our model demonstrates strong performance in shorter videos (i.e., less than $16$ sec), tracking objects becomes challenging in longer setting, with performance (in $\mathcal{J}_{\text{tr}}$-score) dropping from $43$ to $27$ in videos longer than $40$ sec (or containing more than $200$ frames). Despite this, our model outperforms AOT by $\sim 10\%$ in longer video setting (\textit{i.e.}, video length between $34$ and $40$ sec) - confirming the effectiveness of our clip-based memory and matching strategies in long-range scenarios. 

Additionally, the results on small objects (SM) subset is shown in~\cref{suppl:exp:subset_obj}, where we plot $\mathcal{J}_{\text{tr}}$ scores on different object size ranges (as a percentage of the frame size). We observe that performance decreases as object size decreases, dropping from $\sim 65$ to as low as $25$ with objects smaller than $0.2\%$ of the frame size, further confirming the complexity of the task. However, we note that our model consistently outperforms the prior best model, AOT~\cite{yang2021associating}, across all subsets, highlighting the effectiveness of our multi-scale object tracking design.

\paragraph{DAVIS'17 subsets analysis.} In the main submission Table~\textcolor{red}{3c}, we demonstrate that our model achieves competitive performance on a conventional video object segmentation dataset compared to previous work. Table~\ref{tab:exp:davis_subset_2} shows additional results comparing our model with state-of-the-art method (XMem~\cite{cheng2022xmem}) on different DAVIS'17 subsets with respect to object size to analyze performance on small objects. It clearly shows that our model is better equipped to capture small-sized objects on conventional benchmarks, demonstrating the efficacy of our unified multi-scale memory matching and decoding framework. It is worth noting that on subsets of extremely small objects ($< 0.3\%$ and $< 0.05\%$), our model consistently outperforms XMem by approximately $2\%$, showcasing the effectiveness of our design.

\subsection{Additional ablation studies}
\input{tables/suppl_inference_cs}
\paragraph{Memory ablations.}
In the main submission Section~\textcolor{red}{3.4}, we introduced the memory module implemented as a FIFO (first-in-first-out) queue, initially populated with the features of the frame and the ground-truth mask from the initial reference frame. During training, due to computational limitations, we set the memory bank size to 7 and clip-length of 2 for the memory module. However, during inference, when gradients are not computed, we have more GPU memory at our disposal, allowing us to expand the bank size and utilize smaller clip-lengths to include more frames in memory, potentially improving predictions~\cite{press2021train}. Note that we apply linear interpolation on RTE to expand it accordingly for larger bank size. Hence, in this subsection, we delve into the impact of employing different bank sizes and clip lengths for the memory module during inference. The corresponding results are presented in \cref{suppl:tab:test_clip} and \cref{suppl:tab:test_bank}. In \cref{suppl:tab:test_clip}, we observe a performance decrease as clip length increases, corroborating the  observations made by a prior clip-based method PCVOS~\cite{park2022per}. 
This suggests that larger clip lengths might enable tracking global features across lengthy videos, but  it entails less frequent update to the memory possibly compromising the tracking.
On the other hand, the results from testing with different bank sizes are shown in \cref{suppl:tab:test_bank}. We observe a slight increase in performance from increasing the bank size from $7$ to $9$, but decrease in performance with the further increase. This suggests that enabling the model to include more frames in memory during inference results in increase in model's capacity to perform more accurate matching over longer-contexts, but reaches a peak in performance as the model was not trained on larger bank sizes to effectively use them.

\paragraph{Transformation-aware loss ablations.}
\input{tables/suppl_TAR_method}

As mentioned in the main submission Section~\textcolor{red}{3.5}, we compute weights for each frame in the proposed re-weighting objective in different ways. Specifically, we explored three different methods: connected components, center of mass, and masked area. For the masked area, we compute the changes in the foreground area of the binary masks. For the center of mass, we calculate the changes in the relative center of mass. The relative center of mass is calculated as the center of the foreground binary mask, considering the top-left corner of the mask as the origin. Lastly, for the connected components method, we employ {\tt OpenCV} tools to identify the number of distinct groups or regions within the mask. The results of these approaches are presented in \cref{tab:ablation:reweighting_method}.
Notably, simply using the change in the masked area as the weights in the proposed objective yielded the most improvement, resulting in $\sim 2\%$ boost for both $\mathcal{J}^\text{mean}$ and $\mathcal{J}_\text{tr}^\text{mean}$. Based on this observation, we adopted the use of masked area as the default method for computing the weights in our experiments.

Additionally, we note that our transformation-aware re-weighting strategy can be applied to individual or different combinations of the component segmentation losses (DICE~\cite{milletari2016v} and Focal~\cite{lin2017focal}) computed at the frame-level.
We observe in \cref{tab:ablation:reweighting_apply} that applying re-weighting only to the Focal loss yields the best performance. Therefore, in our experiments, we use re-weighting on the focal loss alone as the default setting to achieve optimal performance.

Lastly, in our novel re-weighting loss we introduced a parameter, denoted as $\tau$, in the normalization process. This parameter allows us to control the smoothness of the weights assigned to each frame. In \cref{suppl:fig:ablation_tau}, we empirically assess the effectiveness of different values of $\tau$. We observe that smaller values of $\tau$ result in higher performance. However, when $\tau$ falls below 1.0, the weights become overly sharp, causing a decrease in performance. Note that the larger the value of $\tau$ the smoother the weights become. Based on these results, we used $\tau=1.0$ as our default setting for the re-weighting loss.

\input{figures/ablation_tau}
\subsection{Additional qualitative results}

\paragraph{VOST.}

We conduct a qualitative ablation of our proposed approach in~\cref{suppl:fig:demo_ablation}. When comparing our model's performance (third row) and that without multi-scale (or single-scale) matching (last row), we observe that performing matching on a single-scale allows the model to locate the object of interest. However, it struggles to perform precise segmentation, resulting in false positives.

Additionally, when comparing our model (third row), ours without RTE (fourth row) and ours without re-weighting (fifth row), we notice that the baselines fail to track the object after transformations. This leads to false positives again, e.g., from confusing a wristband for a chili. Conversely, our best model adeptly captures the chili without confusion even after the chili is cut, thereby confirming the effectiveness of our design in video object segmentation, particularly in handling transformations.

\input{figures/suppl_demo_visor}
\paragraph{VISOR.} 
To illustrate our prediction results on VISOR, we provide qualitative results in~\cref{suppl:fig:demo_visor}. In this example, the target video contains 6 objects of interest with different appearance and size. Our model is able to, not only track all objects without confusion, but also capture the objects with tiny size. This demonstrates the effectiveness of our model in challenging scenarios.

\clearpage

\subsection{Additional Visualization and Analysis}
\paragraph{Multi-scale matching}
\input{figures/suppl_demo_attn}
We visualize our multi-scale matching on VOST in~\cref{suppl:fig:demo_attn}. For the top-side results, we select a video with multiple instances. Traditional approaches, which only consider matching on a single scale (i.e., scale 1), may struggle when dealing with objects with similar visual appearances at coarse feature maps. This limitation is evident in the matching attention results at scale 1, where the model might confuse different onions and fail to distinguish between them. In contrast, our multi-scale design, depicted in the scale 2 matching attention results, enables the model to capture subtle visual differences between two onions, focusing more accurately on the correct onion.

For the bottom-side results, we select a video with small instances. In the last attention map results, the model struggles to match the object on the coarse features (scale 1), confusing the object of interest with the human hand and other objects. However, in finer features (scale 2) matching, it successfully attends to the object of interest (the herbs being cut).
These findings illustrate that our multi-scale matching not only allows the model to capture smaller objects but also potentially prevents the model from confusing objects with similar visual appearance.
\clearpage

\input{figures/suppl_demo_reweight}

\paragraph{Transformation-aware re-weighting}

In Section~\textcolor{red}{3.5} in the main submission, we propose the transformation-aware re-weighting to enable the model to place greater emphasis on objects during transformations. In this subsection, we provide examples of training data along with the calculated weights $w^t$ for re-weighting to illustrate how this mechanism operates in practical scenarios. These results are showcased in~\cref{suppl:fig:demo_reweight}. The figure depicts a video sequence where a person starts cutting an eggplant in the second frame, leading to a significant change in the foreground mask's area. Our designed re-weighting strategy calculates weights for each frame based on these observed area changes. The weights, displayed in the first row, highlight the highest peak occurring in the second frame, aligning to the frame with the highest degree in object transformation. Consequently, applying these computed weights to re-weight the loss enables us to focus on frames where objects undergo complex transformations.

\paragraph{Multiplicative Relative Time Encoding}

As outlined in Section~\textcolor{red}{3.2} in the main submission, our multi-scale matching encoder integrates relative-time-encoding (RTE) to discern the significance of each frame within the memory. In this subsection, we present a qualitative demonstration of the learned embeddings in RTE, depicted in~\cref{exp:vis_RTE}. The results showcase the outcomes obtained using element-wise multiplication in Eq.~\textcolor{red}{3}.
The findings in both figures illustrate how our RTE highlights the importance of frames during the matching module. Notably, the results reveal that the matching of the current frame heavily relies on its neighboring frames compared to those further away in the sequence. Furthermore, regardless of the number of frames in the memory, the first couple of frames in the memory always hold significance, as it represents the ground truth mask and its neighbouring frame.
\clearpage

\input{figures/suppl_demo_failure}
\subsection{Limitation and failure cases}
While our method excels in video object segmentation with object shape and appearance transformations, yet it still faces various challenges as follows:
(1) \textit{Object moves out of frame:} In our video demonstration (``video$\_$demo.mp4''), there is a scenario where the object of interest temporarily moves out of the field of view. Our model, like prior methods such as AOT, struggles to successfully track the object upon its return. When the object is out of view, methods relying on matching the current frame with previous frames encounter difficulties in retrieving it due to the lack of the object presence in the frame history.
(2) \textit{Object occlusion with complex transformation}: We illustrate this type of failure in~\cref{suppl:fig:demo_failure}. In this figure, a worker is spreading cement, and our model, along with other methods, can track both the shovel and cement initially. However, once the shovel obstructs the view of the cement in the subsequent frames while the cement undergoes complex transformation, the models lose track of the cement and can only track the shovel. Occlusion events like this, pose challenges for current video object segmentation approaches when objects are undergoing such significant deformations.
These cases present significant challenges for existing video object segmentation methodologies. We leave these directions for future research.

\section{Societal impact}
\label{sec:societal}
Semi-automatic video object segmentation has multiple positive societal impacts as it can be used for a variety of useful applications, \eg robot manipulation and augmented/virtual reality. The ability to track and segment objects in a class agnostic manner can be used in such application areas to enable better human interaction with the environment and improve the user experience. The ability to track under complex transformations is crucial when deploying in the wild in these applications.

However, as with many artificial intelligence algorithms, segmentation and tracking can have negative societal impacts, \eg, through application to target tracking in military systems. To some extent, movements are emerging to limit such applications, \eg pledges on the part of researchers to ban the use of artificial intelligence in weaponry systems. We have participated in signing that pledge and are supporters of its enforcement through international laws.

%% file: tables/suppl_dataset.tex
\begin{table}[t]
\centering
\caption{\textbf{Dataset statistics for VISOR and VOST}. Avg. len. stands for average video length, and Ann. fps denotes annotation frames per second.}  
\resizebox{0.6\columnwidth}{!}{
\begin{tabular}{l|cc|cc|c}
\toprule
\multirow{2}{*}{Dataset} & \multirow{2}{*}{\begin{tabular}[c]{@{}c@{}}Videos\\ (train/val)\end{tabular}} & \multirow{2}{*}{\begin{tabular}[c]{@{}c@{}}Frames\\ (train/val)\end{tabular}} & \multirow{2}{*}{\begin{tabular}[c]{@{}c@{}}Avg. len.\\ (sec)\end{tabular}} & \multirow{2}{*}{\begin{tabular}[c]{@{}c@{}}Ann.\\ fps\end{tabular}} & \multirow{2}{*}{\begin{tabular}[c]{@{}c@{}}Avg. obj size \\ (\% of frame)\end{tabular}} \\
 &  &  &  &  &  \\ \hline
VISOR & 5.3k / 1.2k & 33k / 7.7k & 12.0 & 0.5 & 6.67  \\
VOST & 572 / 70 & 60k / 8k   & 21.2 & 5.0 & 2.57 \\
\bottomrule
\end{tabular}}
\label{suppl:tab:datasets}
\end{table}

%% file: figures/suppl_subset_objectsize.tex
\begin{figure*}[t]
  \centering
  \begin{subfigure}[t]{0.48\textwidth}
  \centering
  \includegraphics[page=5,trim={5 660 353 0}, clip, width=\columnwidth]{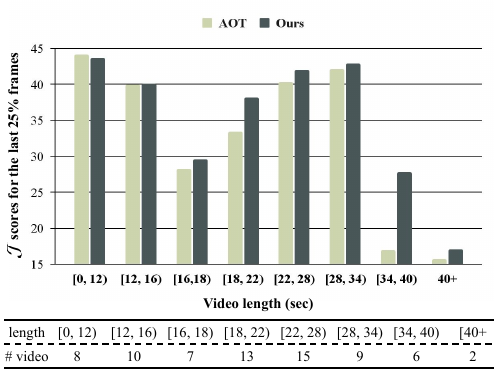}
  \caption{ }
  \label{suppl:exp:subset_video}
  \end{subfigure}
  \begin{subfigure}[t]{0.48\textwidth}
  \centering
  \includegraphics[page=4,trim={5 660 353 0}, clip, width=\columnwidth]{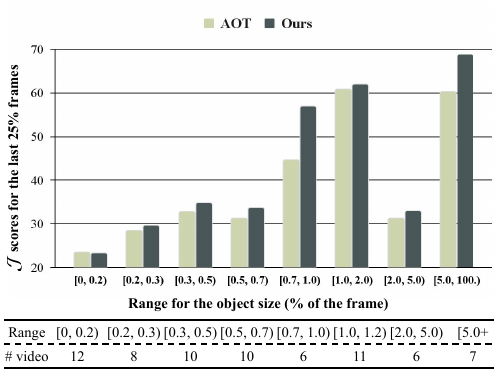}
  \caption{ }
  \label{suppl:exp:subset_obj}
  \end{subfigure}
  \caption{\textbf{Performance breakdown w.r.t. (a) video length and (b) small object size.} We show performance comparison of AOT~\cite{yang2021associating} and our method on subsets of different video lengths and object sizes. (a) We observe that 1) performance decreases in long duration scenarios, demonstrating the complexity in long videos, and 2) our method generally outperforms AOT~\cite{yang2021associating} on long-range subsets, especially upto $10\%$ on videos longer than $34$ secs. (b) We observe that 1) with smaller object size, the performance decreases confirming the complexity of the task, and 2) our method outperforms on all small object subsets compared to AOT~\cite{yang2021associating} demonstrating the effectiveness of our method on small objects (SM). \# video denotes the number of videos.}
\end{figure*}

%% file: tables/suppl_inference_cs.tex
\begin{table*}[t!]
\centering
\caption{\textbf{(a) Performance breakdown on DAVIS'17 val set w.r.t. small object sizes.} \textbf{Performance on VOST dataset w.r.t. (b) clip lengths during inference and (c) bank size during inference.} (a) Note that these models are jointly trained with YouTubeVOS. (b) We observe a peak in distribution of performance w.r.t. clip lengths during inference. This suggests that while larger clip lengths (going from $1$ to $2$) allow for accommodating context from longer time spans, it leads to decrease in performance with further increase ($2+$) as information at granular time-scale is lost. (c) We observe a peak in performance w.r.t. bank sizes. This suggests that, on one hand, increase in bank size leads to increase in model's capacity to perform matching with more number of elements in memory (or longer-context) which translates to increase in performance. But, on the other hand, we observe decrease in performance upon further increase in bank size, suggesting that since the model was not trained on larger bank sizes, it fails to effectively utilize them. Bold number denotes the best performance.}
\begin{subtable}{0.32\linewidth}
\centering
\resizebox{0.98\columnwidth}{!}{
\begin{tabular}{c|c|c|cc}
\toprule
\multicolumn{1}{c|}{\multirow{2}{*}{Approach}} & \multicolumn{3}{c}{Object size} \\ \cline{2-4} 
& $<0.5\%$ & $<0.3\%$ & $<0.05\%$ \\ \midrule
XMem~\cite{cheng2022xmem} & 63.9 & 39.7 & 33.8 \\
\ourmodel (Ours) & \textbf{64.1} & \textbf{43.1} & \textbf{35.9} \\
\bottomrule
\end{tabular}
}
\caption{}
\label{tab:exp:davis_subset_2}
\end{subtable}
\begin{subtable}[t]{.32\linewidth}
\centering
\resizebox{0.90\columnwidth}{!}{
\begin{tabular}{c|cc|cc}
\toprule
Idx & Bank size & Clip length & $\mathcal{J}_{\text{tr}}$ & $\mathcal{J}$ \\ \hline
(1) & 9 & 1 & 35.5 & 47.4 \\
(2) & 9 & 2 & \textbf{37.7} & \textbf{49.3} \\
(3) & 9 & 3 & 32.7 & 44.2 \\
(4) & 9 & 4 & 32.8 & 44.7 \\
\bottomrule
\end{tabular}}
\caption{}
\label{suppl:tab:test_clip}
\end{subtable}
\begin{subtable}[t]{.32\linewidth}
\centering
\resizebox{0.85\columnwidth}{!}{
\begin{tabular}{c|cc|cc}
\toprule
Idx & Bank size & Clip length & $\mathcal{J}_{\text{tr}}$ & $\mathcal{J}$ \\ \hline
(1) & 7 & 2 & 36.2 & 48.2 \\
(2) & 8 & 2 & 36.2 & 48.1 \\
(3) & 9 & 2 & \textbf{37.7} & \textbf{49.3} \\
(4) & 10 & 2 & 36.2 & 48.2 \\
(5) & 11 & 2 & 36.4 & 48.9 \\
(6) & 12 & 2 & 35.4 & 47.8 \\
\bottomrule
\end{tabular}}
\caption{}
\label{suppl:tab:test_bank}
\end{subtable}
\vspace{-5pt}
\vspace{-5pt}
\end{table*}

%% file: tables/suppl_TAR_method.tex
\begin{table*}[t!]
\caption{\textbf{Performance w.r.t. (a) different re-weighting methods used in transformation-aware loss and (b) reweighting combinations of Dice and Focal segmentation loss}. (a) During training, we compute weights (or importance) of each frame while computing total loss using different methods: 1) \textit{Connected components}: weights proportional to relative change in the number of connected components, 2) \textit{Center of mass}: weights proportional to relative change in center of mass, and 3) \textit{Masked area}: weights proportional to relative change in area of foreground objects. We observe best performance when re-weighting our loss with the \textit{Masked Area} method. (b) We observe best performance when reweighting Focal loss (case 2) only, which we use as a default setting for all our experiments.}
\begin{subtable}[t]{.48\linewidth}
\begin{center}
\resizebox{0.95\columnwidth}{!}{
\begin{tabular}{c|c|cc}
\toprule
Method & Re-weighting & $\mathcal{J}_{\text{tr}}$ & $\mathcal{J}$ \\ \hline
Baseline & - & 34.6 & 46.1  \\ 
(1) & Connected components & 35.4 & 46.3 \\
(2) & Center of mass & 34.9 & 45.2   \\
(3) & Masked area & \textbf{36.5} & \textbf{48.2}   \\
\bottomrule
\end{tabular}
}
\end{center}
\caption{}
\label{tab:ablation:reweighting_method}
\end{subtable}
\begin{subtable}[t]{.48\linewidth}
\begin{center}
\resizebox{0.9\columnwidth}{!}{
\begin{tabular}{c|cc|cc}
\toprule
\multirow{2}{*}{Method} & \multicolumn{2}{c|}{Re-weighting} & \multirow{2}{*}{$\mathcal{J}_{\text{tr}}$} & \multirow{2}{*}{$\mathcal{J}$} \\ \cline{2-3}
 & Dice loss & Focal loss &  &  \\ \hline
Baseline &  &  & 34.6 & 46.1 \\ 
(1) & \checkmark &  & 33.2 & 45.4 \\
(2) &  & \checkmark & \textbf{36.5} & \textbf{48.2} \\
(3) & \checkmark & \checkmark & 33.4 & 46.0 \\ 
\bottomrule
\end{tabular}
}
\end{center}
\caption{}
\label{tab:ablation:reweighting_apply}
\end{subtable}
\vspace{-5pt}

\vspace{-5pt}
\end{table*}

%% file: figures/ablation_tau.tex
\begin{figure*}[t]
  \centering
  \begin{subfigure}[t]{0.48\textwidth}
  \includegraphics[page=1,trim={2 0 2 0}, clip, width=\columnwidth]{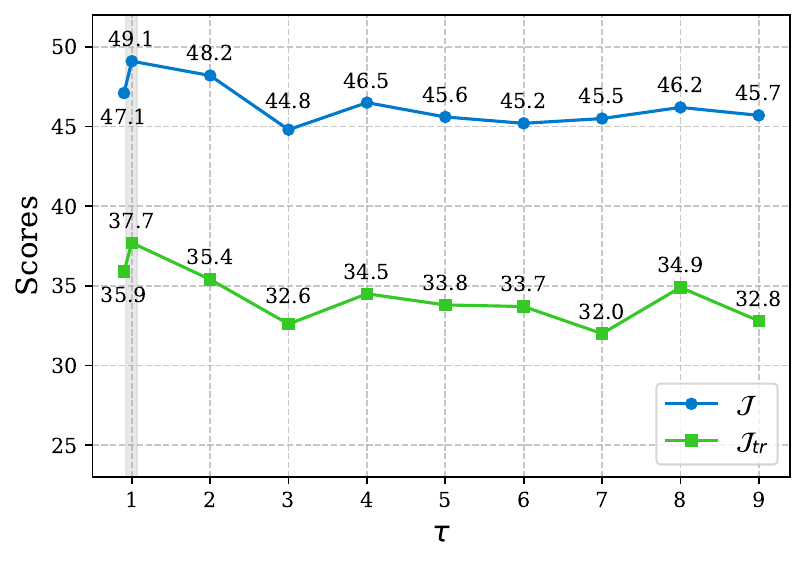}
  \caption{ }
  \label{suppl:fig:ablation_tau}
  \end{subfigure}
  \begin{subfigure}[t]{0.48\textwidth}
  \includegraphics[page=9,trim={0 490 170 0}, clip, width=\columnwidth]{figures/sources/demo.pdf}
  \caption{ }
  \label{suppl:fig:demo_ablation}
  \end{subfigure}
  \caption{\textbf{(a) Performance w.r.t. different $\tau$ used in the transformation-aware reweighting}. We observe peak in performance ($\mathcal{J}_{\text{tr}}$) at $\tau=1.0$, which we use as a default setting for all our experiments. \textbf{(b) Qualitative comparison on VOST for different model ablations.} Best viewed in color; red indicates incorrect predictions.}
\end{figure*}

%% file: figures/suppl_demo_visor.tex
\begin{figure*}[t]
  \centering
  \includegraphics[page=7,trim={0 410 120 0}, clip, width=\columnwidth]{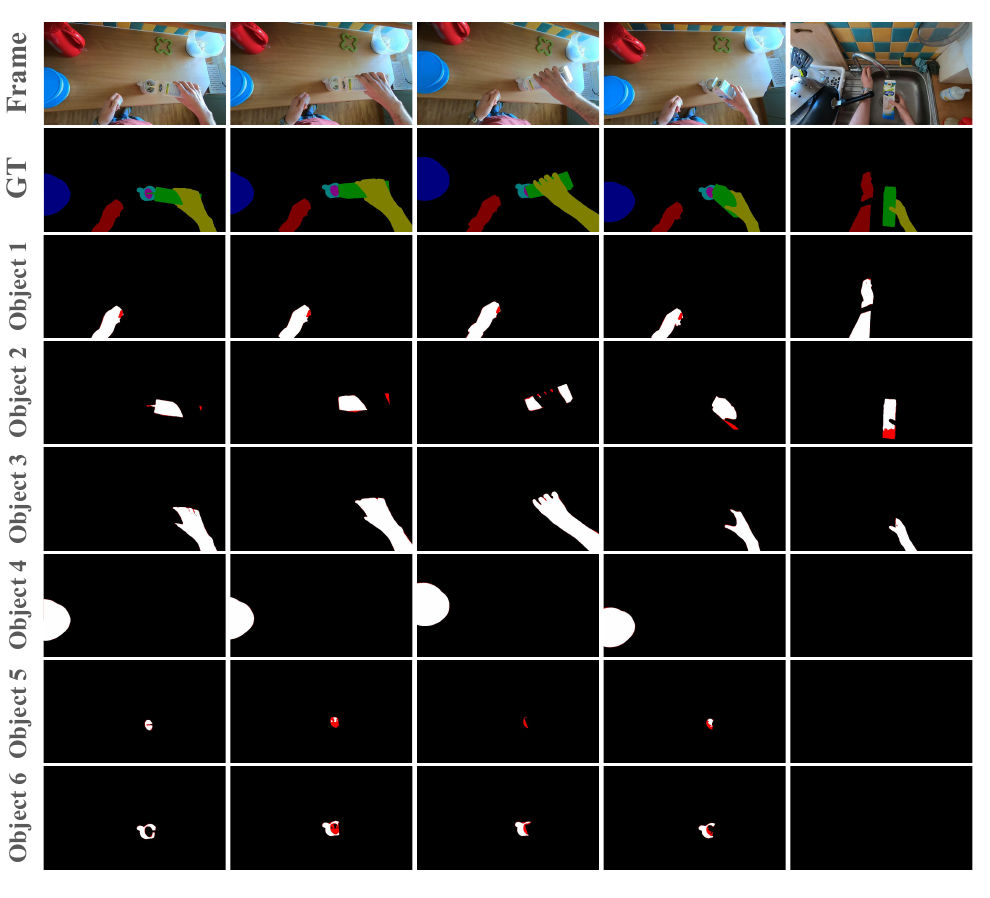}
  \vspace{-0.2in}  
  \caption{\textbf{Qualitative results on VISOR.} Best viewed in color; red indicates incorrect predictions. }
  \vspace{-0.1in}
  \label{suppl:fig:demo_visor}
\end{figure*}

%% file: figures/suppl_demo_attn.tex
\begin{figure*}[t]
  \centering
  \includegraphics[page=6,trim={20 500 20 0}, clip, width=\columnwidth]{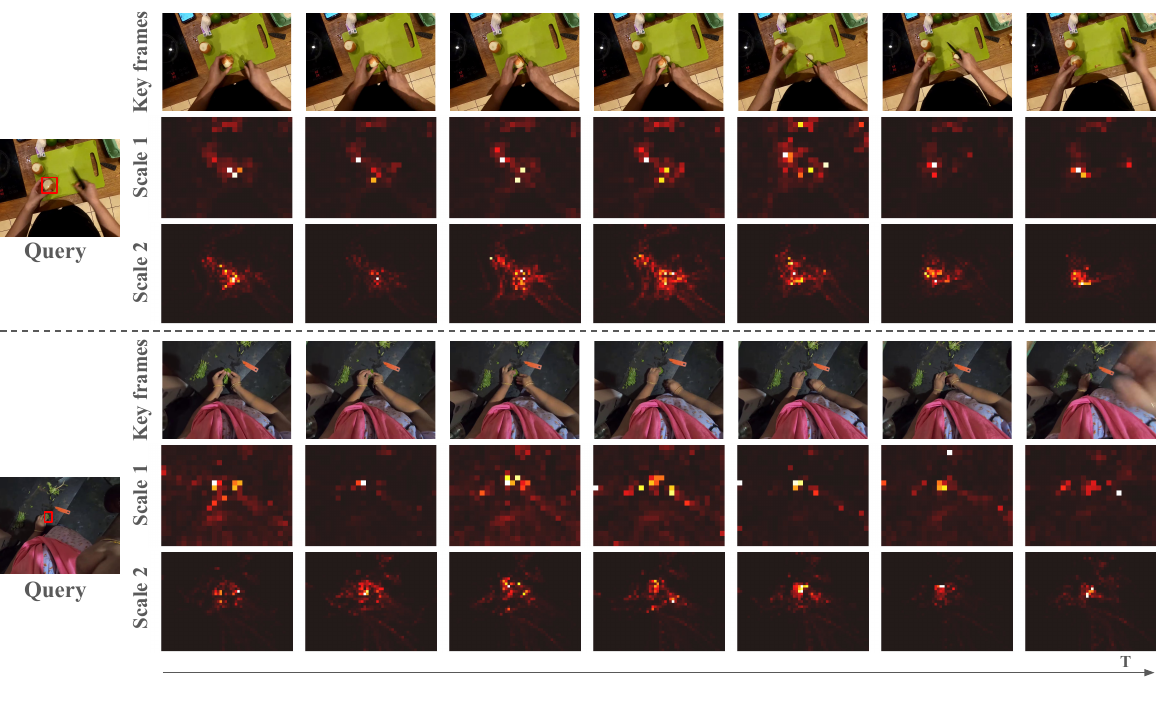}
  \vspace{-0.2in}  
  \caption{\textbf{Visualization of the attention maps} in the multi-scale matching modules. Scale 1 has a resolution of $\frac{1}{32}$ of the input frame, and scale 2 is at $\frac{1}{16}$ resolution. Best viewed in color; lighter colors indicate higher attention scores. The red box in the first frame denotes the object of interest in the query frame.}
  \vspace{-0.1in}
  \label{suppl:fig:demo_attn}
\end{figure*}

%% file: figures/suppl_demo_reweight.tex
\begin{figure}[t]
  \centering
  \begin{subfigure}[t]{0.48\textwidth}
  \includegraphics[page=10,trim={0 620 220 0}, clip, width=\columnwidth]{figures/sources/demo.pdf}
  \caption{}
  \label{suppl:fig:demo_reweight}
  \end{subfigure}
  \begin{subfigure}[t]{0.48\textwidth}
  \includegraphics[page=1,trim={390 28 50 0}, clip, width=0.8\columnwidth]{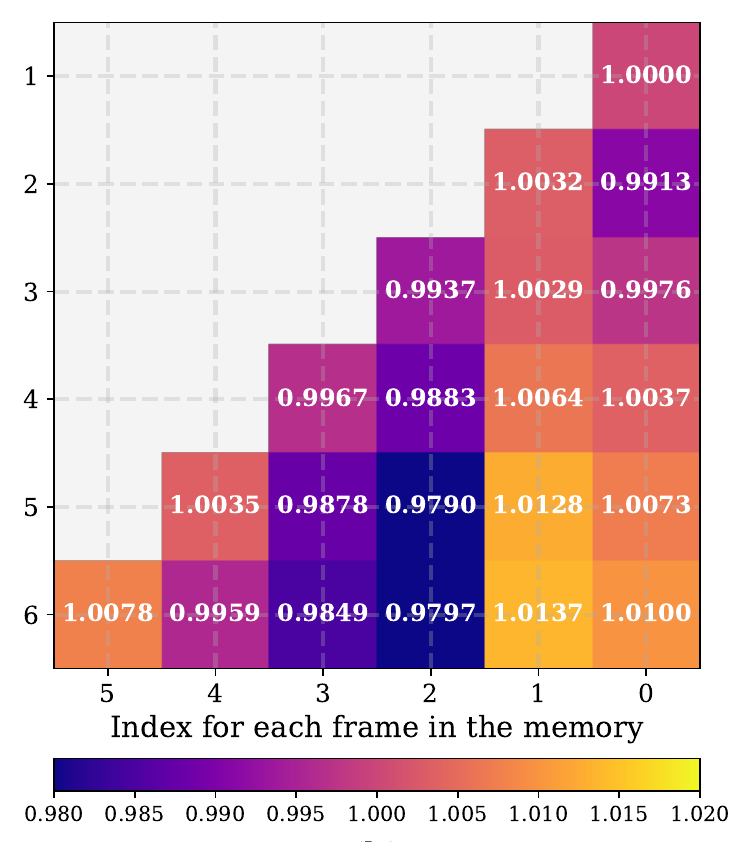}
  \caption{}
  \label{exp:vis_RTE}
  \end{subfigure}
  \caption{\textbf{(a) Visualization of Transformation-aware reweighting.} \textbf{(b) Visualization of RTE.} Value in each grid represents the learned importance score for each frame w.r.t. different number of frames in the memory ($n$). Lighter colors indicate higher scores.}
\end{figure}

%% file: figures/suppl_demo_failure.tex
\begin{figure*}[t]
  \centering
  \includegraphics[page=8,trim={0 595 75 0}, clip, width=\columnwidth]{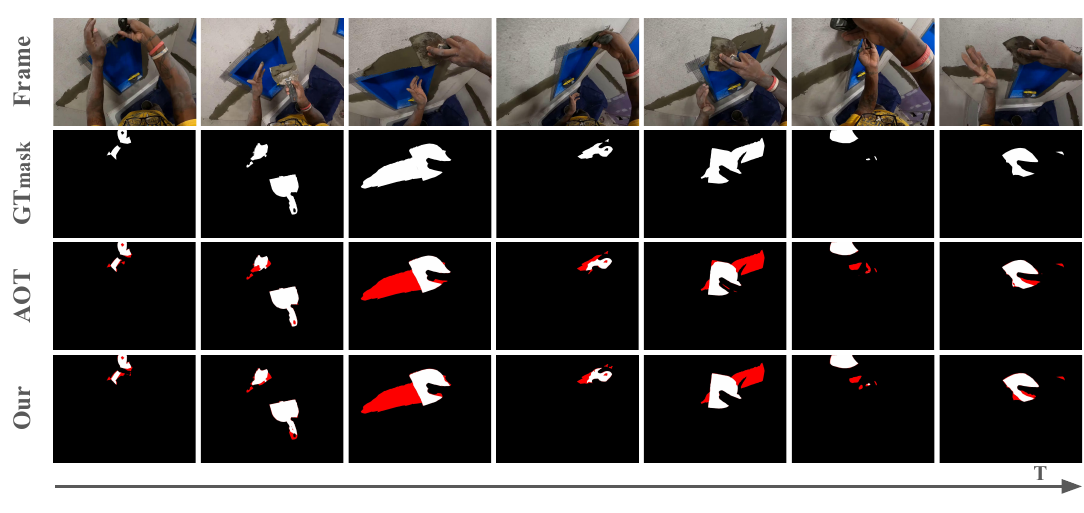}
  \vspace{-0.2in}  
  \caption{\textbf{Failure case.} Best viewed in color; red indicates incorrect predictions. Note that incorrect predictions cover situations where predictions are missing or when objects are wrongly predicted with another object's index, particularly occurring when multiple objects are present in the video. For this figure, both AOT and our model have missed predictions (shown in red color).}
  \vspace{-0.1in}
  \label{suppl:fig:demo_failure}
\end{figure*}

%% file: main.bbl
\begin{thebibliography}{10}
\providecommand{\url}[1]{\texttt{#1}}
\providecommand{\urlprefix}{URL }
\providecommand{\doi}[1]{https://doi.org/#1}

\bibitem{athar2022hodor}
Athar, A., Luiten, J., Hermans, A., Ramanan, D., Leibe, B.: Hodor: High-level object descriptors for object re-segmentation in video learned from static images. In: CVPR (2022)

\bibitem{ba2016layer}
Ba, J.L., Kiros, J.R., Hinton, G.E.: Layer normalization. arXiv preprint arXiv:1607.06450  (2016)

\bibitem{bao2018cnn}
Bao, L., Wu, B., Liu, W.: Cnn in mrf: Video object segmentation via inference in a cnn-based higher-order spatio-temporal mrf. In: CVPR (2018)

\bibitem{caelles2017one}
Caelles, S., Maninis, K.K., Pont-Tuset, J., Leal-Taix{\'e}, L., Cremers, D., Van~Gool, L.: One-shot video object segmentation. In: CVPR (2017)

\bibitem{carion2020end}
Carion, N., Massa, F., Synnaeve, G., Usunier, N., Kirillov, A., Zagoruyko, S.: End-to-end object detection with transformers. In: ECCV (2020)

\bibitem{cheng2022masked}
Cheng, B., Misra, I., Schwing, A.G., Kirillov, A., Girdhar, R.: Masked-attention mask transformer for universal image segmentation. In: CVPR (2022)

\bibitem{cheng2023putting}
Cheng, H.K., Oh, S.W., Price, B., Lee, J.Y., Schwing, A.: Putting the object back into video object segmentation. arXiv preprint arXiv:2310.12982  (2023)

\bibitem{cheng2022xmem}
Cheng, H.K., Schwing, A.G.: Xmem: Long-term video object segmentation with an atkinson-shiffrin memory model. In: ECCV (2022)

\bibitem{cheng2021rethinking}
Cheng, H.K., Tai, Y.W., Tang, C.K.: Rethinking space-time networks with improved memory coverage for efficient video object segmentation (2021)

\bibitem{cheng2014global}
Cheng, M.M., Mitra, N.J., Huang, X., Torr, P.H., Hu, S.M.: Global contrast based salient region detection. TPAMI  (2014)

\bibitem{cho2022tackling}
Cho, S., Lee, H., Lee, M., Park, C., Jang, S., Kim, M., Lee, S.: Tackling background distraction in video object segmentation. In: ECCV (2022)

\bibitem{Damen2022RESCALING}
Damen, D., Doughty, H., Farinella, G.M., , Furnari, A., Ma, J., Kazakos, E., Moltisanti, D., Munro, J., Perrett, T., Price, W., Wray, M.: Rescaling egocentric vision: Collection, pipeline and challenges for epic-kitchens-100. International Journal of Computer Vision (IJCV)  \textbf{130},  33–55 (2022), \url{https://doi.org/10.1007/s11263-021-01531-2}

\bibitem{darkhalil2022epic}
Darkhalil, A., Shan, D., Zhu, B., Ma, J., Kar, A., Higgins, R., Fidler, S., Fouhey, D., Damen, D.: Epic-kitchens visor benchmark: Video segmentations and object relations. In: NeurIPS (2022)

\bibitem{deng2009imagenet}
Deng, J., Dong, W., Socher, R., Li, L.J., Li, K., Fei-Fei, L.: Imagenet: A large-scale hierarchical image database. In: CVPR (2009)

\bibitem{duke2021sstvos}
Duke, B., Ahmed, A., Wolf, C., Aarabi, P., Taylor, G.W.: Sstvos: Sparse spatiotemporal transformers for video object segmentation. In: CVPR (2021)

\bibitem{everingham2010pascal}
Everingham, M., Van~Gool, L., Williams, C.K., Winn, J., Zisserman, A.: The pascal visual object classes (voc) challenge. IJCV  (2010)

\bibitem{feichtenhofer2017spatiotemporal}
Feichtenhofer, C., Pinz, A., Wildes, R.P.: Spatiotemporal multiplier networks for video action recognition. In: Proceedings of the IEEE conference on computer vision and pattern recognition. pp. 4768--4777 (2017)

\bibitem{hariharan2011semantic}
Hariharan, B., Arbel{\'a}ez, P., Bourdev, L., Maji, S., Malik, J.: Semantic contours from inverse detectors. In: ICCV (2011)

\bibitem{he2016deep}
He, K., Zhang, X., Ren, S., Sun, J.: Deep residual learning for image recognition. In: CVPR (2016)

\bibitem{hendrycks2016gaussian}
Hendrycks, D., Gimpel, K.: Gaussian error linear units (gelus). arXiv preprint arXiv:1606.08415  (2016)

\bibitem{hong2023lvos}
Hong, L., Chen, W., Liu, Z., Zhang, W., Guo, P., Chen, Z., Zhang, W.: Lvos: A benchmark for long-term video object segmentation. In: ICCV (2023)

\bibitem{hu2018videomatch}
Hu, Y.T., Huang, J.B., Schwing, A.G.: Videomatch: Matching based video object segmentation. In: ECCV (2018)

\bibitem{jabri2020space}
Jabri, A., Owens, A., Efros, A.: Space-time correspondence as a contrastive random walk. In: NeurIPS (2020)

\bibitem{jampani2017video}
Jampani, V., Gadde, R., Gehler, P.V.: Video propagation networks. In: CVPR (2017)

\bibitem{jang2017online}
Jang, W.D., Kim, C.S.: Online video object segmentation via convolutional trident network. In: CVPR (2017)

\bibitem{karim2023med}
Karim, R., Zhao, H., Wildes, R.P., Siam, M.: Med-vt: Multiscale encoder-decoder video transformer with application to object segmentation. In: Proceedings of the IEEE/CVF Conference on Computer Vision and Pattern Recognition. pp. 6323--6333 (2023)

\bibitem{kim2023universal}
Kim, D., Kim, J., Cho, S., Luo, C., Hong, S.: Universal few-shot learning of dense prediction tasks with visual token matching. In: ICLR (2023)

\bibitem{li2022recurrent}
Li, M., Hu, L., Xiong, Z., Zhang, B., Pan, P., Liu, D.: Recurrent dynamic embedding for video object segmentation. In: CVPR (2022)

\bibitem{liang2020video}
Liang, Y., Li, X., Jafari, N., Chen, J.: Video object segmentation with adaptive feature bank and uncertain-region refinement. In: NeurIPS (2020)

\bibitem{lin2017feature}
Lin, T.Y., Doll{\'a}r, P., Girshick, R., He, K., Hariharan, B., Belongie, S.: Feature pyramid networks for object detection. In: CVPR (2017)

\bibitem{lin2017focal}
Lin, T.Y., Goyal, P., Girshick, R., He, K., Doll{\'a}r, P.: Focal loss for dense object detection. In: ICCV (2017)

\bibitem{lin2014microsoft}
Lin, T.Y., Maire, M., Belongie, S., Hays, J., Perona, P., Ramanan, D., Doll{\'a}r, P., Zitnick, C.L.: Microsoft coco: Common objects in context. In: ECCV (2014)

\bibitem{loshchilov2017decoupled}
Loshchilov, I., Hutter, F.: Decoupled weight decay regularization. arXiv preprint arXiv:1711.05101  (2017)

\bibitem{milletari2016v}
Milletari, F., Navab, N., Ahmadi, S.A.: V-net: Fully convolutional neural networks for volumetric medical image segmentation. In: 3DV (2016)

\bibitem{oh2019video}
Oh, S.W., Lee, J.Y., Xu, N., Kim, S.J.: Video object segmentation using space-time memory networks. In: ICCV (2019)

\bibitem{park2022per}
Park, K., Woo, S., Oh, S.W., Kweon, I.S., Lee, J.Y.: Per-clip video object segmentation. In: CVPR (2022)

\bibitem{perazzi2016benchmark}
Perazzi, F., Pont-Tuset, J., McWilliams, B., Van~Gool, L., Gross, M., Sorkine-Hornung, A.: A benchmark dataset and evaluation methodology for video object segmentation. In: CVPR (2016)

\bibitem{pont20172017}
Pont-Tuset, J., Perazzi, F., Caelles, S., Arbel{\'a}ez, P., Sorkine-Hornung, A., Van~Gool, L.: The 2017 davis challenge on video object segmentation. arXiv preprint arXiv:1704.00675  (2017)

\bibitem{press2021train}
Press, O., Smith, N.A., Lewis, M.: Train short, test long: Attention with linear biases enables input length extrapolation. arXiv preprint arXiv:2108.12409  (2021)

\bibitem{raffel2020exploring}
Raffel, C., Shazeer, N., Roberts, A., Lee, K., Narang, S., Matena, M., Zhou, Y., Li, W., Liu, P.J.: Exploring the limits of transfer learning with a unified text-to-text transformer. JMLR  (2020)

\bibitem{seong2021hierarchical}
Seong, H., Oh, S.W., Lee, J.Y., Lee, S., Lee, S., Kim, E.: Hierarchical memory matching network for video object segmentation. In: Proceedings of the IEEE/CVF International Conference on Computer Vision. pp. 12889--12898 (2021)

\bibitem{shi2015hierarchical}
Shi, J., Yan, Q., Xu, L., Jia, J.: Hierarchical image saliency detection on extended cssd. In: TPAMI (2015)

\bibitem{tokmakov2023breaking}
Tokmakov, P., Li, J., Gaidon, A.: Breaking the" object" in video object segmentation. In: CVPR (2023)

\bibitem{vaswani2017attention}
Vaswani, A., Shazeer, N., Parmar, N., Uszkoreit, J., Jones, L., Gomez, A.N., Kaiser, {\L}., Polosukhin, I.: Attention is all you need. In: NeurIPS (2017)

\bibitem{voigtlaender2019feelvos}
Voigtlaender, P., Chai, Y., Schroff, F., Adam, H., Leibe, B., Chen, L.C.: Feelvos: Fast end-to-end embedding learning for video object segmentation. In: CVPR (2019)

\bibitem{voigtlaender2017online}
Voigtlaender, P., Leibe, B.: Online adaptation of convolutional neural networks for video object segmentation. arXiv preprint arXiv:1706.09364  (2017)

\bibitem{xiao2018monet}
Xiao, H., Feng, J., Lin, G., Liu, Y., Zhang, M.: Monet: Deep motion exploitation for video object segmentation. In: CVPR (2018)

\bibitem{xu2018youtube}
Xu, N., Yang, L., Fan, Y., Yang, J., Yue, D., Liang, Y., Price, B., Cohen, S., Huang, T.: Youtube-vos: Sequence-to-sequence video object segmentation. In: ECCV (2018)

\bibitem{yang2022tubedetr}
Yang, A., Miech, A., Sivic, J., Laptev, I., Schmid, C.: Tubedetr: Spatio-temporal video grounding with transformers. In: CVPR (2022)

\bibitem{yang2018efficient}
Yang, L., Wang, Y., Xiong, X., Yang, J., Katsaggelos, A.K.: Efficient video object segmentation via network modulation. In: CVPR (2018)

\bibitem{yang2020collaborative}
Yang, Z., Wei, Y., Yang, Y.: Collaborative video object segmentation by foreground-background integration. In: ECCV (2020)

\bibitem{yang2021associating}
Yang, Z., Wei, Y., Yang, Y.: Associating objects with transformers for video object segmentation. In: NeurIPS (2021)

\bibitem{yang2021collaborative}
Yang, Z., Wei, Y., Yang, Y.: Collaborative video object segmentation by multi-scale foreground-background integration. TPAMI  (2021)

\bibitem{yang2022decoupling}
Yang, Z., Yang, Y.: Decoupling features in hierarchical propagation for video object segmentation. In: NeurIPS (2022)

\bibitem{yu2023video}
Yu, J., Li, X., Zhao, X., Zhang, H., Wang, Y.X.: Video state-changing object segmentation. In: ICCV (2023)

\bibitem{zhou2022survey}
Zhou, T., Porikli, F., Crandall, D.J., Van~Gool, L., Wang, W.: A survey on deep learning technique for video segmentation. TPAMI  (2022)

\end{thebibliography}
